\documentclass[journal]{IEEEtran}
\usepackage{amsthm}
\newtheorem{definition}{Definition}

\usepackage{cite}
\usepackage{amsmath,amssymb,amsfonts}
\usepackage{algorithmic}
\usepackage{algorithm}
\usepackage{graphicx}
\usepackage{textcomp}
\usepackage{xcolor}
\usepackage{multirow}
\usepackage{hyperref}
\def\BibTeX{{\rm B\kern-.05em{\sc i\kern-.025em b}\kern-.08em
		T\kern-.1667em\lower.7ex\hbox{E}\kern-.125emX}}

\newtheorem{theorem}{Theorem}
\newtheorem{lemma}[theorem]{Lemma}

\ifCLASSINFOpdf
\else
\fi
\begin{document}
\IEEEoverridecommandlockouts
\title{FedFeat+: A Robust Federated Learning Framework Through Federated Aggregation and Differentially Private Feature-Based Classifier Retraining \vspace{2mm}}
%
%
%
\author{Mrityunjoy~Gain,~\IEEEmembership{Student Member,~IEEE,}
	    Kitae~Kim,
        Avi Deb~Raha,~\IEEEmembership{Student Member,~IEEE,}
        Apurba~Adhikary,
        Eui-Nam~Huh,
        Zhu~Han,~\IEEEmembership{Fellow,~IEEE,}
        and~Choong Seon~Hong,~\IEEEmembership{Fellow,~IEEE}
\thanks{Mrityunjoy Gain is with the Department of Artificial Intelligence, School of Computing,
	Kyung Hee University, Yongin 17104, Republic of Korea. (e-mail: gain@khu.ac.kr).
	
	Kitae Kim, Avi Deb Raha, Eui-Nam Huh, and Choong Seon Hong are with the Department of Computer Science and Engineering, School of Computing, Kyung Hee University, Yongin 17104, Republic of Korea. (e-mail: glideslope@khu.ac.kr; avi@khu.ac.kr; johnhuh@khu.ac.kr; cshong@khu.ac.kr).
	
	Apurba Adhikary is with the Department of Computer Science and Engineering, School of Computing, Kyung Hee University, Yongin-si 17104, Republic of Korea, and also with the Department of Information and Communication Engineering, Noakhali Science and Technology University, Noakhali-3814, Bangladesh (e-mail: apurba@khu.ac.kr).
	
	Zhu Han is with the Electrical and Computer Engineering Department,
	University of Houston, Houston, TX 77004 (email: hanzhu22@gmail.com).
	
	Corresponding author: Choong Seon Hong (e-mail: cshong@khu.ac.kr).
	}
	\vspace{-2mm}
}

%
%

\markboth{Preprint Submitted to IEEE Journal for Peer Review}%
{Shell \MakeLowercase{\textit{et al.}}: Bare Demo of IEEEtran.cls for IEEE Journals}
%



\maketitle

\begin{abstract}
Federated learning faces significant hurdles when dealing with complex, imbalanced, and non-independent and identically distributed (non-IID) client data, thus often leading to suboptimal model accuracy. Traditional aggregation methods struggle to effectively capture the diversity of data across clients. Without a holistic view of the collective data from all clients, these models often lack robust generalization. To address those challenges, in this paper, we propose the FedFeat+ framework, which distinctively separates feature extraction from classification. We develop a two-tiered model training process: following local training, clients transmit their weights and some features extracted from the feature extractor from the final local epochs to the server. The server aggregates these models using the FedAvg method and subsequently retrains the global classifier utilizing the shared features. The classifier retraining process enhances the model's understanding of the holistic view of the data distribution, ensuring better generalization across diverse datasets. This improved generalization enables the classifier to adaptively influence the feature extractor during subsequent local training epochs. We establish a balance between enhancing model accuracy and safeguarding individual privacy through the implementation of differential privacy mechanisms. By incorporating noise into the feature vectors shared with the server, we ensure that sensitive data remains confidential. We present a comprehensive convergence analysis, along with theoretical reasoning regarding performance enhancement and privacy preservation. We validate our approach through empirical evaluations conducted on benchmark datasets, including CIFAR-10, CIFAR-100, MNIST, and FMNIST, achieving high accuracy while adhering to stringent privacy guarantees. The experimental results demonstrate that the FedFeat+ framework, despite using only a lightweight two-layer CNN classifier, outperforms the FedAvg method in both IID and non-IID scenarios, achieving accuracy improvements ranging from 3.92\% to 12.34\% across CIFAR-10, CIFAR-100, and Fashion-MNIST datasets.
\end{abstract}

\begin{IEEEkeywords}
FedFeat+, classifier retrain, privacy preserve, differential privacy, feature sharing.
\end{IEEEkeywords}

%
\IEEEpeerreviewmaketitle

\section{Introduction}
\IEEEPARstart{T}he Internet of Things (IoT) has enabled billions of interconnected devices to generate massive amounts of decentralized data across diverse environments. This explosion of distributed data brings both opportunities and challenges, particularly in terms of privacy and computational limitations \cite{iot1}. 
Federated learning (FL) is a decentralized approach that allows multiple edge clients to collaboratively train a shared model while keeping their data local \cite{raha_dg}. This decentralized method is increasingly important as privacy concerns and regulations grow, thereby emphasizing the need to protect sensitive information. By enabling models to learn directly from data stored on individual devices, FL enhances privacy and reduces the risk of data breaches \cite{girum_1, girum_2}. Its relevance is especially pronounced in sectors like healthcare, finance, and IoT, where safeguarding user confidentiality is crucial while still benefiting from collective insights \cite{mcmahan2017communication}.

Current research in FL is advancing rapidly, driven by the increasing need for privacy-preserving machine learning solutions. Key areas of focus include improving communication efficiency, enhancing model accuracy, and developing robust algorithms that can handle non-IID (independent and identically distributed) data across clients. Researchers are exploring new methods for optimizing federated training processes, such as adaptive aggregation techniques and improved noise addition strategies to ensure privacy without sacrificing performance. Recent studies also emphasize the importance of theoretical guarantees for convergence and privacy, establishing a solid foundation for deploying FL in real-world scenarios. However, FL often faces significant challenges in achieving high accuracy due to the non-IID nature of client data, particularly when data distributions are extremely imbalanced \cite{xiao_1}. This imbalance distribution the model's ability to generalize effectively, as certain classes may be underrepresented, leading to suboptimal performance. The aggregated model fails to adequately capture the diversity of the underlying data, compromising the efficacy of FL systems \cite{xiao_2}. Therefore, there is a pressing need for robust strategies that address these imbalances and enhance model performance by ensuring better generalization across varying client distributions and capturing the holistic data distribution.

To address this issue, we propose the FedFeat+ framework, which fundamentally rethinks the FL paradigm by separating the feature extraction and classification processes. After local training, clients send the model weights and some extracted features from the last local epoch to the central server. We utilize the FedAvg \cite{mcmahan2017communication} aggregation method to effectively combine the weights shared by clients. The feature-sharing mechanism ensures that the server receives a comprehensive representation of the data, thereby enhancing the model's ability to learn from the variety of distributions present across different clients. Then the server aggregates the features of the clients and retrains the global classifier by the features. By focusing on essential features that capture the distinct patterns of each client’s data, our FedFeat+ framework significantly mitigates the challenges posed by data imbalance, ultimately leading to improved model accuracy and performance. However, feature sharing poses a significant threat of privacy leakage. To address this, we employ a differential privacy mechanism, incorporating noise into the features shared among clients to ensure sensitive information about individual data points is sufficiently masked. This mechanism allows us to maintain the confidentiality of personal data while enabling effective feature aggregation. By carefully calibrating the amount of noise based on the sensitivity of the features and the desired privacy budget, we achieve a balance between privacy protection and model performance. Our theoretical and empirical evaluations demonstrate that the FedFeat+ framework enhances model accuracy while maintaining robust privacy guarantees which is beneficial for IoT applications. The contributions of our research encompass three pivotal aspects:

\begin{itemize}
	\item We propose FedFeat+, a novel FL framework that enhances standard FL by introducing a classifier retraining stage at the server. After local training, clients share differentially private feature representations extracted during their final local epochs. The server uses these representations to retrain the global classifier, capturing a more holistic view of the data distribution and enhancing model performance and robustness.
	\item For safeguarding individual privacy, we integrate a differential privacy mechanism that effectively masks the output features generated by local clients. This approach ensures that sensitive information remains confidential while still allowing for valuable insights to be extracted from the data.
	\item The experimental results demonstrate that the FedFeat+ framework, utilizing only a two-layer classifier backbone for CNN, outperforms the FedAvg method in both IID and non-IID scenarios, improving accuracy over FedAvg for CIFAR-10 by up to 3.92\% in IID and 7.62\% in Non-IID, for CIFAR-100 by 9.47\% in IID and 12.34\% in Non-IID, and for Fashion-MNIST by 8.94\% in IID and 11.75\% in Non-IID.
\end{itemize}

The remainder of this paper is organized as follows. Section \ref{LR} discusses the literature review. In Section \ref{Method}, we illustrated our proposed FedFeat+ method. We present the convergence analysis of our proposed FedFeat+ method in Sections \ref{CA}. Privacy assurance of our proposed FedFeat+ method is illustrated in Section \ref{PA}. We present the theoretical proof of performance improvement for our FedFeat+ method in Section \ref{PE}. Section \ref{Exp} presents the experiments, and finally Section \ref{Con} concludes the paper.

\section{Literature Review} \label{LR}
In this section, we provide an overview of FL, Split Learning (SL), and Federated Split Learning (FSL), discussing their background and baseline research.
\begin{figure*}[t]
	\centerline{\includegraphics[width=\textwidth]{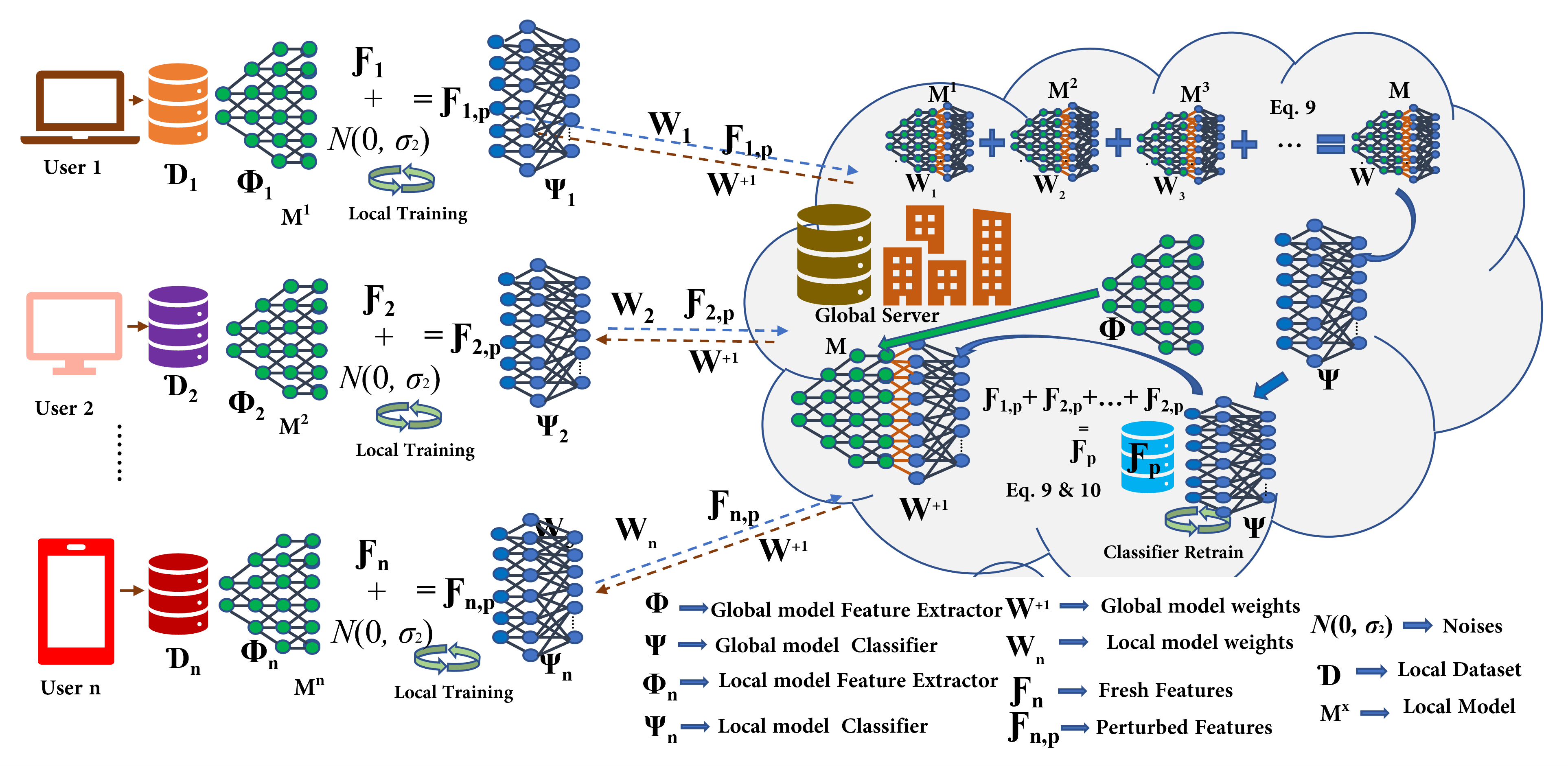}}
	\caption{System overview of proposed FedFeat+ framework.}
	\label{system-model}
\end{figure*}
In 2017, McMahan et al. \cite{mcmahan2017communication} introduced FedAvg, a foundational framework in FL that enables collaborative model training across distributed devices while safeguarding user data privacy through decentralized updates. FedAvg facilitates local model training on individual devices, followed by aggregation of these updates on a central server. To address the inherent challenges of FL, subsequent research has focused on enhancing its robustness and efficiency. Li et al. \cite{fedprox} proposed incorporating a penalty term to stabilize training, particularly in scenarios where client data distributions exhibit significant variations. Building upon this, Wang et al. \cite{wang2020federated} introduced FedMA, an adaptive aggregation mechanism that weights client updates based on their individual model performance, thereby improving the overall training process. Addressing the issue of "client drift," where local models diverge significantly from the global model, Karimireddy et al. \cite{karimireddy2020scaffold} developed SCAFFOLD, a novel approach that employs corrective updates to align local and global model parameters, leading to improved model accuracy and generalization. Furthermore, researchers have explored innovative techniques to enhance the adaptability and robustness of FL. MOON \cite{moon} leverages contrastive learning to improve model alignment across clients with diverse data distributions. Additionally, FedPer \cite{perso} introduces personalization layers to tailor models to the unique characteristics of individual client data, further enhancing model performance. Raha et al. \cite{raha_icact} propose a hybrid FL framework with edge devices, regional edge servers (RES), and cloud servers (CS). Lightweight models run on the edge, while complex cases are offloaded to RES using uncertainty-based semantic offloading for improved decision-making. To improve efficiency and scalability, Sattler et al. \cite{clustered} proposed clustered FL (CFL), which divides clients into groups to reduce communication overhead and accelerate the training process. Reddi et al. \cite{opt} investigated the crucial role of optimization algorithms in FL, exploring adaptations of existing optimizers and developing novel methods to accelerate training convergence and improve overall robustness. Li et al. \cite{iofl} propose IOFL which uses client feedback to guide server-side optimization, improving accuracy. Zhong et al. \cite{wvfl} propose WVFL which ensures secure, verifiable aggregation with encrypted values. Shi et al. \cite{sam} SAM reduces communication by selectively aggregating models. A study on wireless IoT \cite{chen} enhances efficiency through model reuse and resource optimization. Song et al. \cite{g_fl} propose Guard-FL which improves robustness using UMAP-based feature clustering. 

In 2018, the concept of split learning emerged as a promising approach to address the challenges of privacy and computational limitations in distributed machine learning. This innovative paradigm divides a neural network across multiple entities, enabling collaborative training while minimizing the need to share raw data. Vepakomma et al. \cite{vepakomma2018split} pioneered this concept by introducing a distributed framework where clients train a portion of the model locally and share only intermediate activations with a central server, significantly reducing the risk of data exposure. Building upon this foundation, Gupta et al. \cite{gupta} expanded split learning to facilitate collaborative deep learning across multiple data sources, enabling data owners to collaborate without compromising data locality. Recognizing the persistent privacy concerns, Pathak et al. \cite{pathak2020fedsplit} developed FedSplit, a novel approach that leverages the alternating direction method of multipliers (ADMM) to efficiently coordinate local updates and server-based aggregation, enhancing both privacy and training efficiency. To further enhance privacy and security, Thapa et al. \cite{thapa2022splitfed} proposed SplitFed, a framework that partitions the model into distinct components: clients perform initial feature extraction, while the server handles classification. This approach minimizes the amount of sensitive information exchanged between clients and the server. Addressing the challenge of data heterogeneity across clients, Shen et al. \cite{ringsfl} introduced RingSFL, a novel architecture that organizes clients into a ring structure, enabling collaborative training while accommodating variations in data distributions. Li et al. \cite{Li_2022_CVPR} developed ResSFL, a robust split learning framework that incorporates model inversion resistance to mitigate the risk of privacy breaches. By incorporating an attacker-aware feature extractor, ResSFL enhances the privacy of client data by making it more difficult for adversaries to reconstruct sensitive information from shared activations.

However, little work has considered the scenario of classifier retraining and the performance degradation that arises in scenarios characterized by extreme non-IID distributions and imbalanced data, which remain critical issues in the field. In this paper, we propose a novel framework, FedFeat+, which focuses on retraining the classifier based on shared features for better accuracy. To ensure privacy, we incorporate a differential privacy mechanism, safeguarding sensitive information during the retraining process.

\section{Methodology}\label{Method}

In this section, we present the detailed procedure and mechanisms of our proposed FedFeat+ framework. The overview of our proposed FedFeat+ framework is shown in Fig. \ref{system-model}. By separating feature extraction from classification, the FeadFeat+ method allows clients to share some features that capture the diversity of their local datasets. Following local training, clients transmit their model weights and extracted features to the central server. Below, we outline the step-by-step procedure, highlighting the key operational mechanisms that contribute to improved performance in FL.

\textbf{Model Initialization}
We begin the FL process by initializing the model parameters for both the feature extractor and the classifier using a Gaussian distribution as follows:
\begin{equation}
	\mathcal{M}_0 = \{ \mathbf{w}_i \sim \mathcal{N}(\mu, \sigma^2) \; | \; \forall \mathbf{w}_{i} \in \mathcal{W} \},
	\label{init_model}
\end{equation}
where $ \mathbf{w}_i $ denotes the model parameters, $ \mu $ is the mean of the distribution (typically set to 0), $ \sigma^2 $ is the variance (a small positive constant), and $ \mathcal{W} $ is the set of all weights in the model. Here, $ \mathcal{M}_0 $ represents the initial set of model parameters, which are randomly initialized before any training begins. The model is comprises of a feature extractor ($\Phi$) and a classifier ($\Psi$).

\textbf{Local Training at Client $i$:} Each client $i$ computes features from its local dataset using a feature extraction function $\Phi$ as follows:
\begin{equation}
	\mathcal{F}_i = \Phi(X_i),
	\label{fe}
\end{equation}
where $\mathcal{F}_i$ denotes the features extracted from the input data $X_i$ at client $i$. The feature extractor processes the input to generate features that will be used by the local classifier for the learning process. To ensure privacy, we employ a differential privacy mechanism. The clients add Gaussian noise to the extracted features as follows:
\begin{equation}
	\mathcal{F}_{i,\text{perturbed}} = \mathcal{F}_i + ( N(0, \sigma^2) || \text{Lap}(0, \lambda) ),
	\label{gs}
\end{equation}
where $ \mathcal{F}_{i,\text{perturbed}} $ represents the perturbed features after adding Gaussian noise / Laplacian noise to mitigate privacy risks. This noise, modeled as $ N(0, \sigma^2) $ / $Lap(0, \lambda)$, ensures that individual data points cannot be easily identified \cite{data_retrival}. The sensitivity $ \Delta $ of the feature extraction function $ \mathcal{F} $ is defined as follows:
\begin{equation}
	\Delta = \max_{x, x'} \left| \mathcal{F}(x) - \mathcal{F}(x') \right|,
	\label{sen}
\end{equation}
where $ \Delta $ quantifies the maximum change in the output of the function $ \mathcal{F} $ resulting from a single change in the input data. It reflects how sensitive the feature extraction function is to individual data points. The noise variance $ \sigma^2 $ is calculated based on the privacy budget $ \epsilon $ as follows:
\begin{equation}
	\sigma^2 = \frac{2 \Delta^2 \log(1/\delta)}{\epsilon^2}, \lambda = \frac{\Delta}{\epsilon}.
	\label{var}
\end{equation}

This equation determines the variance of the Gaussian / Laplacian noise added for differential privacy. The parameter $ \epsilon $ controls the trade-off between privacy and utility, $\Delta$ is the sensitivity, and $ \delta $ represents the failure probability of the privacy mechanism.
The client then passes the perturbed features through the classifier ($\Psi$) and trains both the feature extractor and classifier by minimizing the loss function $\mathcal{L}$, which compares the predicted labels with the actual labels $Y_i$ as follows:
\begin{align}
	\mathcal{M}_i^{(t)} & = \arg\min_{\mathcal{M}} \mathcal{L}(Y_i, \hat{Y}_i) \\
	& = \arg\min_{\mathcal{M}} \sum_{(X_j, Y_j) \in C_i} \mathcal{L}(Y_i,\Psi(\Phi(X_i)),
	\label{cf}
\end{align}
where, the term $\mathcal{L}$ is the local loss function, which evaluates how well the model's predictions $\hat{Y}_i$ match the actual labels $Y_j$ for each data point $(X_j, Y_j)$ in the client's dataset $C_i$. The objective is to minimize this loss by updating the feature extractor($\Phi$) and classifier's($\Psi$) parameters using local data.

\textbf{Server-Side Aggregation (FedAvg):} After each client completes local training, they send their local model and some perturbed features from the last local epochs to the server. The server aggregates both the parameters of the feature extractor $\Phi$ and classifier $\Psi$ to build the global model using FedAvg as follows:
\begin{equation}
	\Phi^{(t+1)} = \frac{1}{N} \sum_{i=1}^{N} \frac{|\mathcal{D}_i|}{|\mathcal{D}|}\Phi_i^{(t)}, \quad \Psi^{(t+1)} = \frac{1}{N} \sum_{i=1}^{N} \frac{|\mathcal{D}_i|}{|\mathcal{D}|}\Psi_i^{(t)},
	\label{fedavg}
\end{equation}
where $\Phi^{(t+1)}$ and $\Psi^{(t+1)}$ denote the global parameters for the feature extractor and the classifier at global round $t+1$, respectively. Similarly, $\Phi_i^{(t)}$ and $\Psi_i^{(t)}$ represent the local parameters for the feature extractor and classifier from client $i$ at round $t$. $N$ is the total number of clients participating in the round. $|\mathcal{D}_i|$ is the size of client $i$'s dataset, and $|\mathcal{D}|$ is the total size of all datasets.


\textbf{Features Aggregation:} 
The shared perturbed features from the local clients can be represented as follows:
\begin{equation}
	\mathcal{F} = \{ \mathcal{F}_{1,\text{perturbed}}, \mathcal{F}_{2,\text{perturbed}}, \dots, \mathcal{F}_{N,\text{perturbed}} \},
	\label{agg}
\end{equation}
where $ \mathcal{F} $ is the set of all perturbed features shared by the clients with the central server. The set includes the perturbed feature vectors $ \mathcal{F}_{1,\text{perturbed}}, \mathcal{F}_{2,\text{perturbed}}, \dots, \mathcal{F}_{N,\text{perturbed}} $, ensuring privacy during transmission.
The central server concatenates the perturbed features received from all clients as follows:
\begin{equation}  
	\mathcal{F}_p = \bigoplus_{i=1}^{N} \mathcal{F}_{i,\text{perturbed}},
	\label{concat}
\end{equation}  
where $ \oplus $ represents the concatenation operator, and the resulting vector $ \mathcal{F}_p \in \mathbb{R}^{\sum_{i=1}^{N} d_i} $, ensuring that individual client data remains obscured due to the added noise during feature sharing.

\textbf{Classifier Retraining:} Once the server has aggregated the features, it retrains the classifier by minimizing the loss function over the aggregated features as follows:
\begin{align}
	\Psi(t+1) & = \arg \min_{\Psi} \mathcal{L}(Y, {\mathcal{F}_p}; \Psi) \\
	& = \arg \min_{\Psi} \sum_{(X_j, Y_j) \in {\mathcal{F}_p}} \ell({\mathcal{F}_p}, Y_j; \Psi),
	\label{cr}
\end{align}
where $\Psi(t+1)$ represents the updated classifier parameters after retraining, where $\mathcal{L}(Y, {\mathcal{F}_p}; \Psi)$ is the loss function calculated over the aggregated features ${\mathcal{F}_p}$ and the corresponding labels $Y$. This step ensures that the server’s classifier improves its predictions based on the collective client data, while still preserving privacy. The feature extractor remains unchanged this time.


\textbf{Model Redistribution:} 
After retraining the classifier, the server redistributes the updated parameters $\mathcal{M}_{t+1}$ to all participating clients for the next training round. This ensures that the clients continue to improve their local models based on the updated global knowledge.

\textbf{Repeat Process:} 
This entire process is repeated for $T$ global rounds, enabling continual learning and model refinement across multiple iterations while maintaining privacy at each stage. The complete procedure of the proposed FeadFeat+ method is shown in Algorithm \ref{algorithm}. In the following sections, we analyze the theoretical properties of FedFeat+, including its convergence behavior, privacy guarantees, and generalization performance, followed by a complexity analysis and supporting experimental results.

\begin{algorithm}[!t]
	\caption{Algorithm for FedFeat+}
	\label{algorithm}
	\begin{algorithmic}[1]
		\STATE \textbf{Input:} Number of clients $ C $, Number of global rounds $ R $, Model parameters $ \mathcal{M} $ (initialized randomly), Privacy budget $ \epsilon $, Noise scale $ \sigma $ for differential privacy.
		\STATE \textbf{Output:} Trained model parameters $ \mathcal{M}^* $
		\FOR{each global round $ r = 1 $ to $ R $}
		\STATE \textbf{Client Selection:} Select a subset of clients $ S \subseteq C $.
		\FOR{each client $ c \in S $}
		\STATE Receive global model parameters $ \mathcal{M} $.
		\STATE Split the model into feature extractor $ \Phi $ and classifier $ \Psi $.
		\STATE Extract features $ F_c $ from local dataset $ D_c $ by ($\ref{fe}$)
		\STATE Add differential privacy noise to the features by ($\ref{gs}$) 
		\STATE Pass the perturbed features to the classifier and train both feature extractor and classifier on local dataset $ D_c $ to get updated model parameters by ($\ref{cf}$)
		\ENDFOR
		
		\STATE \textbf{Model and Feature Upload:} Send updated model parameters $ \mathcal{M} $ and perturbed features of the last local epochs $ F'_c $ to the server.
		\STATE \textbf{Aggregation:} Aggregate received model parameters by ($\ref{agg}$)
		\STATE \textbf{Retrain Classifier:} Aggregate features from all clients by ($\ref{concat}$)
		\STATE Retrain classifier $ \Psi $ on aggregated features by ($\ref{cr}$)
		\STATE \textbf{End of Rounds:} Return the final model parameters: $\mathcal{M}^* = \mathcal{M}_{\text{final}}\Rightarrow(\Phi_{\text{new}}, \Psi_{\text{final}})$
		\ENDFOR
	\end{algorithmic}
\end{algorithm}
\section{Convergence Analysis for FeadFeat+ Framework} \label{CA}
The convergence of the proposed FeadFeat+ framework is a crucial aspect that ensures efficient model training across clients. This analysis focuses on the loss function characteristics, learning rate dynamics, impact of noise, and classifier retraining.

\subsection{Convergence of FedAvg}
Given the global objective function $ f(\mathbf{w}) $, and assuming that each client performs stochastic gradient descent (SGD) on its local data, the convergence behavior of the FedAvg algorithm is well-established in the literature (e.g., \cite{mcmahan2017communication, fedprox}). Under appropriate conditions for the learning rates, number of communication rounds, and gradient updates, the convergence bound is expressed as follows:
\begin{equation}
	\|\mathbf{w}^{(t+1)} - \mathbf{w}^*\|_2 \leq \gamma \|\mathbf{w}^{(t)} - \mathbf{w}^*\|_2,
\end{equation}
where $ \mathbf{w}^* $ is the global optimum, and $ \gamma $ is a constant factor depending on the learning rate and other parameters such as the number of clients $ N $, the number of local epochs $ E $, and the smoothness of the objective function. To analyze the convergence behavior of FedFeat+, we begin by outlining key assumptions and presenting supporting lemmas. We begin by analyzing the convergence behavior of FedAvg under feature perturbation, which we formalize through the following lemma and its supporting assumptions.

\begin{lemma}[Convergence of FedAvg with Feature Perturbation]
 Under the feature perturbation mechanism (where clients add differential privacy noise to their features), the global model in FedAvg converges to the optimal model with a bounded error, with a convergence rate similar to that of standard FedAvg.
 
 \textbf{Assumptions:} Before proceeding with the proof, we outline an assumption under which the convergence holds. The global objective function is convex and smooth with a Lipschitz continuous gradient, the features $ \mathbf{z}_i $ for client $ i $ are perturbed by noise $ \mathbf{n}_i $, where $ \mathbf{n}_i $ is either Gaussian noise $ \mathcal{N}(0, \sigma^2) $ or Laplace noise $ \text{Lap}(0, b) $, the server aggregates the model updates using FedAvg and retrains the classifier on the perturbed features, and the learning rate is sufficiently small such that the global model converges.
  
 After $ t+1 $ communication rounds, the difference between the global model and the optimal model $ \mathbf{w}^* $ is bounded as:
 \[
 \|\mathbf{w}^{(t+1)} - \mathbf{w}^*\|_2 \leq \gamma' \|\mathbf{w}^{(t)} - \mathbf{w}^*\|_2,
 \]
 where $ \gamma' $ is a constant that depends on the number of local epochs, number of clients, learning rate, and noise level $ \sigma $ or $ b $ (depending on whether Gaussian or Laplace noise is used).
\end{lemma}
We now proceed to prove the lemma by showing that, under Lipschitz continuity, the addition of small feature perturbations leads to bounded deviations in classifier output, thus preserving the convergence behavior of FedAvg.
\begin{proof}	
	We want to demonstrate that adding small perturbations to the features does not significantly affect the classifier's output.	By the Lipschitz continuity property of the classifier $c$, the change in the classifier's output is bounded by the perturbation magnitude in the features. Specifically, for any two inputs $z_i$ and $\tilde{z}_i = z_i + n_i$, according to \cite{lc} we have:
	\begin{equation}
		\| c(z_i) - c(\tilde{z}_i) \| \leq L \| z_i - \tilde{z}_i \|,
		\label{LC1}
	\end{equation}
	where $L$ is the Lipschitz constant of the classifier. Since the perturbation is $z_i - \tilde{z}_i = n_i$, this becomes:
	\begin{equation}
	\| c(z_i) - c(\tilde{z}_i) \| \leq L \| n_i \|.
		\label{LC2}
	\end{equation}
	
	Since $n_i$ represents the added noise, its norm is controlled by the noise distribution (e.g., Gaussian or Laplace). This ensures that the output difference is bounded by $L \| n_i \|$. If $n_i \sim \mathcal{N}(0, \sigma^2)$ (Gaussian Noise), then the expected magnitude of the perturbation (i.e., the expected norm of $n_i$) based on \cite{bishop} is given by:
	\begin{equation}
	\mathbb{E}[\| n_i \|] = \sigma \cdot \sqrt{d},
		\label{GN}
	\end{equation}
	where $d$ is the dimensionality of the feature vector $z_i$, and $\sigma$ is the standard deviation of the Gaussian noise. This ensures that the perturbation's effect is proportional to $\sigma$ and the feature dimensionality. If $n_i \sim \text{Laplace}(0, b)$ (Laplace Noise), the expected norm is proportional to $b$, the scale parameter of the Laplace distribution. Based on \cite{bishop} we can express:
	\begin{equation}
	\mathbb{E}[\| n_i \|] \sim b \cdot \sqrt{d}.
		\label{LN}
	\end{equation}
	
	This provides a similar bounded effect for Laplace noise. By controlling $\sigma$ or $b$, the perturbation's effect can be made arbitrarily small while maintaining privacy.
	Under the assumptions stated in the lemma, the convergence rate of FedAvg, without perturbation, is typically given as:
	\begin{equation}
		\| w^* - w^{\text{global}}_T \| \leq O\left(\frac{1}{T}\right),
	\end{equation}
	where $w^*$ is the optimal model weights, and $w^{\text{global}}_T$ is the global model weights after $T$ rounds. In the presence of perturbation, the noise does not significantly alter the model's convergence rate. The noise is small and bounded as shown in (\ref{LC1}, \ref{LC2}, \ref{GN} and \ref{LN}), and thus does not significantly affect the optimization process. The error introduced by perturbation is bounded, leading to a convergence rate that remains the same as in standard FedAvg. Therefore, the global model $w_T$ converges at the same rate as without perturbation:
	\begin{equation}
		\| w^* - w^{\text{global}}_T \| \leq O\left(\frac{1}{T}\right).
	\end{equation}
	
	The key insight here is that while the noise perturbation may affect the exact model parameters, it does not significantly degrade the classifier's ability to generalize. The classifier, having seen the perturbed features from all clients during retraining, can effectively learn the underlying data patterns. This is because the perturbations are small and bounded by the noise distribution, and the perturbation's impact diminishes over time as the model converges.
	The classifier, after retraining on the aggregated perturbed features, maintains good generalization performance while ensuring that the individual data points remain obscured by the noise.
\end{proof}
Next, we analyze the role of classifier retraining in accelerating convergence. This relationship is formally captured in the following lemma.
\begin{lemma}[Convergence Rate Improvement with Classifier Retraining]
	Let $ w^* $ denote the optimal global model weights, and let $ w^{(t)} $ denote the model weights after $ t $ global aggregation rounds. In the proposed method, the server retrains the classifier using aggregated features after each global round, where the local clients perturb their features with differential privacy noise and share the perturbed features with the server. The convergence rate of the global model improves due to the retraining mechanism.
	Then, the expected squared deviation between the model weights $ w^{(t+1)} $ after retraining and the optimal model $ w^* $ follows the bound:
	\[
	\mathbb{E}[\|w^{(t+1)} - w^*\|^2] \leq O\left( \frac{1}{T^\alpha} \right),
	\]
	where $ \alpha > 1  $, and $ T $ is the number of global rounds. This improved convergence rate is due to the fact that retraining on aggregated features reduces the gradient variance, which leads to faster alignment with the optimal model.
\end{lemma}
We now proceed to prove this lemma by analyzing how retraining on perturbed aggregated features influences convergence behavior.
\begin{proof}
	In the proposed method, before sharing the model weights with the server, the clients perturb their features using differential privacy noise. Let the feature vectors of client $ i $ at global round $ t $ be $ z_i^{(t)} $, and the perturbed feature vector after adding noise $ n_i^{(t)} $ be denoted as $ \tilde{z}_i^{(t)} = z_i^{(t)} + n_i^{(t)} $, where $ n_i^{(t)} $ is a noise term drawn from either a Gaussian or Laplace distribution.
	The key difference between the proposed method and standard FedAvg is the retraining phase. When the server retrains the global model using the aggregated perturbed features, the model gains exposure to a broader and more diverse distribution of features, which improves the alignment with the optimal model $ w^* $. By reducing the gradient variance, the retraining phase ensures that the global model is updated in a more stable manner, thereby accelerating convergence. After aggregation and retraining, the global model $ w^{(t+1)} $ is expected to converge faster towards the optimal solution $ w^* $. Let the deviation between the global model and the optimal model be $ \| w^{(t+1)} - w^* \| $. For standard FedAvg, the convergence rate is proportional to $ O\left(\frac{1}{T}\right) $, but due to the retraining step, the global model $ w^{(t+1)} $ experiences a faster convergence rate. Specifically, the deviation can be bounded as:
	\begin{equation}
		\mathbb{E}\left[\| w^{(t+1)} - w^* \|^2\right] \leq O\left( \frac{1}{T^\alpha} \right),
	\end{equation}
	where $ \alpha > 1 $ depending on the specific perturbation noise and retraining method used. This shows that the convergence rate improves from $ O\left(\frac{1}{T}\right) $ in FedAvg to $ O\left(\frac{1}{T^\alpha}\right) $ in the proposed method. The reason behind this acceleration is that retraining allows the model to benefit from a broader feature distribution. This broader exposure enables the global model to better generalize, leading to reduced bias and variance in the gradient updates. The improved alignment with the optimal model during retraining ensures that the model converges more quickly compared to standard FedAvg.
	
	The retraining phase after aggregation enables the global model to improve its generalization capability by aligning with the optimal solution $ w^* $ faster. This results in a convergence rate improvement, where the expected deviation between the global model and the optimal model scales as $ O\left(\frac{1}{T^\alpha}\right) $ instead of $ O\left(\frac{1}{T}\right) $, where $\alpha > 1  $. Thus, the proposed method with feature perturbation and retraining leads to faster convergence and improved generalization compared to the standard FedAvg approach.
\end{proof}

\section{Privacy Assurance in our FeadFeat+ Framework} \label{PA}
To demonstrate that our FeadFeat+ framework ensures privacy, we leverage the principles of differential privacy alongside the mechanisms used in the feature-sharing and model training process. We structure the proof into several key components. To assess the privacy guarantees of FedFeat+, we begin by introducing essential definitions that form the foundation of our analysis.

\begin{definition}[Feature-Level Differential Privacy (FL-DP)]
A mechanism is considered feature-level differentially private if the addition of controlled noise to the output features ensures that the sensitivity of the mechanism is bounded, thereby preserving privacy. Formally, a mechanism \( \mathcal{M} \) satisfies FL-DP if, for any two input features \( z \) and \( z' \) such that \( \| z - z' \| \leq \Delta \), and for all subsets \( S \) of possible outputs, the following holds based on \cite{dp}:  
\[
\text{Pr}[\mathcal{M}(z) \in S] \leq e^{\epsilon} \cdot \text{Pr}[\mathcal{M}(z') \in S],
\]
where, \( \Delta \) is the sensitivity of the feature transformation, \( \epsilon \) is the privacy budget, and \( \mathcal{M}(z) = z + n \), with \( n \) being noise sampled from a suitable distribution (e.g., Gaussian or Laplacian) calibrated based on \( \Delta \).
\end{definition}

\begin{definition}
\textbf{Privacy Loss:} Privacy loss measures how much information about the original data can be inferred from the perturbed output. A lower privacy loss implies stronger privacy preservation. It is defined as the logarithmic difference between the probabilities of an event $ S $ occurring in the outputs of the mechanism applied to features $ z $ and $ z' $ based on \cite{dp}:
\[
\Delta \mathcal{L} = \left|\text{log} \left(\frac{\text{Pr}[\mathcal{M}(z) \in S]}{\text{Pr}[\mathcal{M}(z') \in S]}\right)\right|,
\]
where $ z $ represents the original features and $ z' $ epresents the perturbed features, with the perturbation being applied to the features to obscure the original data. The perturbation makes the output less dependent on the original values, thereby affecting the privacy loss. 
\end{definition}

\begin{definition}
	\textbf{Reconstruction Difficulty:} Reconstruction difficulty refers to the challenge or impossibility of recovering the original data (e.g., an image) from the perturbed features. This difficulty increases with the amount of noise added to the features. Specifically, the more significant the noise, the less likely it is that an adversary can successfully reconstruct the original data.
\end{definition}
To evaluate the effectiveness of noise in preserving data privacy, we analyze how feature perturbation impacts the ability to reconstruct original inputs. This relationship is formalized in the following lemma.
\begin{lemma}[Perturbed Features Ensure Reconstruction Difficulty]
	Adding noise to the features extracted by the feature extractor ensures that reconstructing the original data from the perturbed features becomes computationally infeasible or highly inaccurate.
\end{lemma}
To support the above lemma, we provide a proof demonstrating how noise addition increases reconstruction difficulty.
\begin{proof}
	Let $\mathbf{z}$ represent the original feature vector extracted from the data, and $\tilde{\mathbf{z}} = \mathbf{z} + \mathbf{n}$, where $\mathbf{n}$ is noise sampled from either a Laplace or Gaussian distribution. To reconstruct the original data $\mathbf{x}$, an adversary typically uses a model $g$ to minimize the reconstruction error:
	\begin{equation}
		\min_{g} \mathbb{E}[\| \mathbf{x} - g(\tilde{\mathbf{z}}) \|^2].
	\end{equation}
	
	The success of reconstruction depends on how closely $\tilde{\mathbf{z}}$ aligns with $\mathbf{z}$, which is disrupted by noise $\mathbf{n}$. The noise $\mathbf{n}$ has a distribution with mean $0$ and variance $\sigma^2$. For Gaussian noise: $\mathbf{n} \sim \mathcal{N}(0, \sigma^2 \mathbf{I}),$ and for Laplace noise: $	\mathbf{n} \sim \text{Laplace}(0, b),$ where $b$ is the scale parameter.
	The perturbed features $\tilde{\mathbf{z}}$ now contain noise that is independent of the original data. Thus, the adversary faces two challenges: 1) Estimating $\mathbf{z}$ accurately from $\tilde{\mathbf{z}}$, and 2) Inferring $\mathbf{x}$ from the noisy $\mathbf{z}$, given the added noise. Based on \cite{data_retrival}, the expected error in reconstructing $\mathbf{z}$ from $\tilde{\mathbf{z}}$ is proportional to the noise variance:
	\begin{equation}
		\mathbb{E}[\| \mathbf{z} - \tilde{\mathbf{z}} \|^2] = \mathbb{E}[\| \mathbf{n} \|^2] = d \cdot \sigma^2,
	\end{equation}
	where $d$ is the dimensionality of $\mathbf{z}$. As $\sigma^2$ (or $b^2$) increases, the reconstruction error grows linearly with the noise scale. For the adversary to recover $\mathbf{x}$ from $\tilde{\mathbf{z}}$, they need:
	\begin{equation}
		g(\tilde{\mathbf{z}}) \approx f_{\text{extractor}}^{-1}(\mathbf{z}),
	\end{equation}
	where $f_{\text{extractor}}$ is the feature extractor. However, the inverse mapping $f_{\text{extractor}}^{-1}$ is non-trivial due to model complexity, and noise $\mathbf{n}$ adds ambiguity, making multiple reconstructions equally probable, reducing the certainty of recovering $\mathbf{x}$. Based on the study \cite{data_retrival}, the probability of reconstructing the exact $\mathbf{x}$ from $\tilde{\mathbf{z}}$ decays exponentially with the noise scale:
	\begin{equation}
		P(\mathbf{x} \mid \tilde{\mathbf{z}}) \propto \exp\left(-\frac{\|\mathbf{n}\|}{b}\right) \quad \text{(Laplace noise)},
	\end{equation}
	\begin{equation}
		P(\mathbf{x} \mid \tilde{\mathbf{z}}) \propto \exp\left(-\frac{\|\mathbf{n}\|^2}{2\sigma^2}\right) \quad \text{(Gaussian noise)}.
	\end{equation}
	
Therefore, by adding noise to the extracted features, the reconstruction error increases significantly. The adversary’s ability to recover the original data $\mathbf{x}$ from the noisy features $\tilde{\mathbf{z}}$ becomes computationally infeasible as the noise level increases. Thus, privacy is preserved.
\end{proof}
Next, we analyze the robustness of the classifier component in FedFeat+ under feature perturbations. Specifically, we show that minor noise added for privacy does not significantly impact the classifier's predictive performance.
\begin{lemma}[Classifier Can Maintain Robustness Under Small Perturbation]  
	Small perturbations applied to the features $ \mathbf{z}_i $ do not significantly affect the output distribution from the classifier $ c(\mathbf{z}_i) $. Consequently, the classifier can effectively learn the original data patterns despite the perturbation.
\end{lemma}
To support this lemma, we demonstrate how the classifier's output remains stable under bounded feature perturbations, leveraging the Lipschitz continuity of the classifier function.
\begin{proof}
	Let the features extracted from the feature extractor be $ \mathbf{z}_i $ for client $ i $, and let $ \tilde{\mathbf{z}}_i $ represent the perturbed features after adding differential privacy noise $ \mathbf{n}_i $, such that $\tilde{\mathbf{z}}_i = \mathbf{z}_i + \mathbf{n}_i,$, where $ \mathbf{n}_i \sim \mathcal{N}(0, \sigma^2) $ or $ \mathbf{n}_i \sim \text{Laplace}(0, b) $, depending on the noise distribution used. The classifier $ c(\cdot) $ maps the feature space to the output probability distribution over labels:
	\begin{equation}
		p(y \mid \mathbf{z}_i) = c(\mathbf{z}_i), \quad p(y \mid \tilde{\mathbf{z}}_i) = c(\tilde{\mathbf{z}}_i).
	\end{equation}

	Using the Lipschitz continuity of the classifier:
	\begin{equation}
		\| c(\mathbf{z}_i) - c(\tilde{\mathbf{z}}_i) \| \leq L \cdot \| \mathbf{z}_i - \tilde{\mathbf{z}}_i \| ,
	\end{equation}
	where $ L $ is the Lipschitz constant. Substituting $ \tilde{\mathbf{z}}_i = \mathbf{z}_i + \mathbf{n}_i $, we get:
	\begin{equation}
		\| c(\mathbf{z}_i) - c(\tilde{\mathbf{z}}_i) \| \leq L \cdot \| \mathbf{n}_i \|.
	\end{equation}
	
	For Gaussian noise $ \mathbf{n}_i \sim \mathcal{N}(0, \sigma^2) $, the expected norm of the noise is:
	\begin{equation}
		\mathbb{E}[\| \mathbf{n}_i \|] = d \cdot \sigma,
	\end{equation}
	where $ d $ is the dimensionality of $ \mathbf{z}_i $. For Laplace noise $ \mathbf{n}_i \sim \text{Laplace}(0, b) $, the expected norm is proportional to $ b $, the scale parameter.	By choosing $ \sigma $ or $ b $ appropriately (small enough to maintain privacy while large enough to preserve utility), the change in output probabilities $ \| c(\mathbf{z}_i) - c(\tilde{\mathbf{z}}_i) \| $ can be made arbitrarily small:
	\begin{equation}
		\| c(\mathbf{z}_i) - c(\tilde{\mathbf{z}}_i) \| \leq L \cdot \mathbb{E}[\| \mathbf{n}_i \|].
	\end{equation}
	
	Since the perturbation-induced changes are bounded, the classifier can learn the underlying patterns in the data even with perturbed inputs. This ensures that the addition of noise does not hinder the classifier’s ability to generalize effectively over the original data distribution. Therefore the classifier is robust to small perturbations introduced by the FL-DP mechanism, enabling it to learn from the perturbed features without significant loss of performance.
\end{proof}

\section{Theoretical Proof of Performance Improvement}\label{PE}
The effectiveness of FL can be greatly improved by separating feature extraction and classifier retraining, with the latter performed on the server side. In this section, we demonstrate why our proposed method outperforms FedAvg in terms of both accuracy and convergence, providing evidence of its superiority in handling FL tasks. We consider
$ N $ is the number of clients, $ \mathcal{D}_i $ is the local dataset at client $ i $, $ \mathbf{w}_{\text{global}} $ is the aggregated model parameters after FedAvg, $ F(\mathbf{w}) $ is the loss function for the global model, $ \mathbf{w}_i $ is the local model parameters at client $ i $, $ \mathbf{Z} = \{ \mathbf{z}_i \}_{i=1}^{N} $ is the aggregated features from all clients, where $ \mathbf{z}_i = f_{\text{feat}}(\mathcal{D}_i) $. $ \mathbf{w}_{\text{class}} $ is the classifier parameters. $ \mathbf{w}_{\text{feat}} $ is the feature extractor parameters. $ L $ is the Lipschitz constant of the gradients, $ L' $ is Lipschitz constant for the feature representation, $ F^* $ is the optimal loss, and $ \mathcal{L}_{\text{classifier}} $ is the loss function specific to the classifier.

\emph{Step 1}: Federated Averaging (FedAvg): The global model parameters are updated using the FedAvg method:

\begin{equation}
	\mathbf{w}_{\text{global}} = \sum_{i=1}^{N} \frac{|\mathcal{D}_i|}{|\mathcal{D}|} \mathbf{w}_{i},
	\label{}
\end{equation}
where $ |\mathcal{D}| = \sum_{i=1}^{N} |\mathcal{D}_i| $, is essentially the weighted average of the local models, where the weight is the fraction of the client’s dataset size $|\mathcal{D}_i|$ relative to the total dataset size $|\mathcal{D}|$. Here, $ \mathbf{w}_{\text{global}} $ represents the aggregated global model parameters. Each client $ i $ contributes its local model parameters $ \mathbf{w}_i $ proportionally to the size of its local dataset $ |\mathcal{D}_i| $, relative to the total size of all datasets $ |\mathcal{D}| $. That's means the weights $\frac{|\mathcal{D}_i|}{|\mathcal{D}|}$ ensure that clients with larger datasets contribute more to the global model. This weighted aggregation ensures fair contribution from all clients.

\emph{Step 2}: Classifier Retraining: The classifier is retrained using aggregated features:
\begin{equation}
	\mathbf{w}_{\text{class}} = \arg \min_{\mathbf{w}} \mathcal{L}_{\text{classifier}}(\mathbf{w}; \mathbf{Z}).
	\label{}
\end{equation}
This equation shows the optimization problem for retraining the classifier. Here, $ \mathbf{w}_{\text{class}} $ represents the classifier's parameters, which are updated by minimizing the classifier loss $ \mathcal{L}_{\text{classifier}} $. The loss function is dependent on the aggregated feature set $ \mathbf{Z} $, which contains feature representations extracted by all clients. By retraining the classifier on these aggregated features, we improve the model's performance and decision boundary.

\emph{Step 3}: Performance Comparison: The loss function for the FedAvg algorithm can be bounded as follows (\cite{fedavg_conver}):  

\begin{equation}
	F(\mathbf{w}_{\text{global}}) \leq F^* + \frac{L}{2} \cdot \left( \frac{1}{N} \sum_{i=1}^{N} \|\mathbf{w}_i - \mathbf{w}^*\|^2 \right),
\end{equation}
where $ F(\mathbf{w}_{\text{global}}) $ is the loss of the global model after aggregation, $ F^* $ is the optimal global loss, achieved when model parameters align perfectly with $ \mathbf{w}^* $, $ L $ is the Lipschitz constant, representing the smoothness of the gradient of $ F $, $ \mathbf{w}_i $ is the model parameters of client $ i $ after local training, and $ \mathbf{w}^* $ is the optimal global model parameters that minimize $ F $.
The inequality provides an upper bound on the loss of the global model, $ F(\mathbf{w}_{\text{global}}) $, by accounting for the theoretical best-case loss ($ F^* $) when all models perfectly align with the global objective, and a penalty term $\frac{L}{2} \cdot \left( \frac{1}{N} \sum_{i=1}^{N} \|\mathbf{w}_i - \mathbf{w}^*\|^2 \right),$ which depends on the average squared deviation between the local models $ \mathbf{w}_i $ and the global optimum $ \mathbf{w}^* $. If the local models are close to the global optimum (i.e., $ \|\mathbf{w}_i - \mathbf{w}^*\|^2 $ is small), the penalty term will also be small. Conversely, if the local models deviate significantly from $ \mathbf{w}^* $, the penalty term becomes larger, leading to a greater gap between $ F(\mathbf{w}_{\text{global}}) $ and $ F^* $.
From the above discussion we find two key insights:
\begin{enumerate}
	\item \textbf{Impact of Heterogeneity:} When client data is non-IID, the local models $ \mathbf{w}_i $ can vary significantly, increasing $ \|\mathbf{w}_i - \mathbf{w}^*\|^2 $. This results in a larger penalty term, slowing down convergence and reducing global model performance.
	\item \textbf{Alignment with Global Objective:} The performance of FedAvg is influenced not only by the quality of local updates but also by how well the local models $ \mathbf{w}_i $ align with the global objective $ \mathbf{w}^* $.
\end{enumerate}

In non-IID scenarios, local models are trained on different data distributions, causing significant variation in $ \mathbf{w}_i $ across clients. Aggregating these diverse models leads to a global model that is farther from $ \mathbf{w}^* $, increasing the penalty term. This, in turn, widens the gap between $ F(\mathbf{w}_{\text{global}}) $ and $ F^* $, resulting in slower convergence and suboptimal performance. To formalize the generalization benefit of FedFeat+, we present the following lemma that analyzes the impact of server-side classifier retraining on performance across heterogeneous client data.
\begin{lemma} [ Server-Side Classifier Retraining Improves Classifier's Generalization]
	By retraining the global classifier on aggregated features $ \mathbf{Z} $ collected from all clients, the classifier achieves better generalization across the global data distribution, even under non-IID settings.  
\end{lemma} 
We now provide a formal analysis to support the above lemma, highlighting how the amount of shared feature information contributes to improved generalization.
\begin{proof}
	Let the local model $ \mathbf{w}_i $ be trained on client $ i $'s data, and let $ \mathbf{Z}_i $ represent the features extracted from the last local epochs by the feature extractor of client $ i $. The aggregated feature representation is $	\mathbf{Z} = \bigcup_{i=1}^N \mathbf{Z}_i$. Now, we retrain the global classifier $ \mathbf{w}_{\text{class}}^{\text{global}} $ using the aggregated features $ \mathbf{Z} $. The classifier loss function is given by:
	\begin{equation}
		F_{\text{class}}(\mathbf{w}_{\text{class}}^{\text{global}}) = \frac{1}{|D_{\text{global}}|} \sum_{(x, y) \in D_{\text{global}}} \mathcal{L}(\mathbf{w}_{\text{class}}^{\text{global}}, x, y),
	\end{equation}
	where $ D_{\text{global}} = \bigcup_{i=1}^N D_i $ represents the aggregated global dataset, consisting of data from all clients, and $ \mathcal{L} $ is the loss function for classification (e.g., cross-entropy). By retraining the classifier with aggregated features, the model learns from a more diverse set of data, which approximates the global data distribution. This aggregated dataset $ D_{\text{global}} $ captures the heterogeneity of the clients' local data distributions, which are typically non-IID. In standard FedAvg, the global model is formed by averaging the local models $\mathbf{w}_{\text{global}} = \frac{1}{N} \sum_{i=1}^N \mathbf{w}_i$.	However, the aggregation step alone does not fully take into account the diversity in the data distribution, which can cause a performance drop, especially in non-IID settings. The key improvement in our proposal is the retraining of the classifier using the aggregated features, which introduces the following benefit:
	\begin{itemize}
		\item The retrained global classifier $ \mathbf{w}_{\text{class}}^{\text{global}} $ is exposed to a more complete data distribution (i.e., $ D_{\text{global}} $).
		\item The loss function $ F_{\text{class}}(\mathbf{w}_{\text{class}}^{\text{global}}) $ minimizes the discrepancy between the global classifier and the features that represent the collective data from all clients.
	\end{itemize}
	
	Thus, the global classifier $ \mathbf{w}_{\text{class}}^{\text{global}} $ is better suited to generalize across the clients' local distributions, reducing the impact of local data heterogeneity. Let's consider the difference in loss between the global model and the local models. Let $ F_{\text{class}}(\mathbf{w}_{\text{class}}^*) $ denote the optimal loss for the global classifier, and let $ F_{\text{class}}(\mathbf{w}_{\text{class}}^{\text{global}}) $ represent the loss after retraining.	The retrained global model loss is:
	\begin{equation}
		F_{\text{class}}(\mathbf{w}_{\text{class}}^{\text{global}}) \leq F_{\text{class}}(\mathbf{w}_{\text{class}}^*) + \frac{L}{2} \|\mathbf{Z} - \mathbf{Z}^*\|^2,
	\end{equation}
	where $ \mathbf{Z}^* $ is the optimal aggregated feature representation, and $ L $ is a constant related to the smoothness of the feature extractor. By retraining the global classifier using the aggregated features, the model is exposed to a broader and more representative view of the data, allowing it to generalize better. This process improves the global model's performance in comparison to local models trained independently. As the model is retrained iteratively using the aggregated features from all clients, the global classifier approaches a more optimal state that generalizes well across the clients' heterogeneous data distributions. Thus, the retraining of the classifier using aggregated features leads to improved generalization across the global data distribution.
\end{proof}
We then examine how a generalized global classifier influences the quality of feature extraction during local training, which we formalize in the following lemma.
\begin{lemma} [ Improved Generalized Classifier Enhances Feature Extractor Quality]
	A more generalized global classifier improves feature extraction during subsequent local training rounds, leading to better alignment of local models with the global objective.    
\end{lemma} 
To support this lemma, we next analyze how the improved classifier gradients influence the training dynamics of the feature extractor at each client.
\begin{proof}
  Once the global classifier $ c_{\text{global}} $ has been retrained using aggregated features, it will typically perform better because it has seen a broader range of data from all clients, and thus, its loss should decrease $F_{\text{global}}^{\text{new}} < F_{\text{global}}^{\text{previous}}$. The retrained global classifier is able to generalize better because it now has access to a wider variety of data. Once the global classifier $ c_{\text{global}}^{\text{new}} $ is retrained and more generalized, it can now be used as a target for the local feature extractors during the next local training round. The local feature extractor is responsible for generating the feature vector $ \mathbf{z}_i $ from each client's data. During training, the feature extractor tries to minimize the loss with respect to the global classifier's predictions. The new loss function for the local feature extractor becomes:
  \begin{equation}
  	\mathcal{L}_i^{\text{new}} = L(\mathbf{z}_i, c_{\text{global}}^{\text{new}}(\mathbf{z}_i)).
  \end{equation}
  
  This shows that each local feature extractor is now adjusting its parameters to make its feature representations $ \mathbf{z}_i $ more compatible with the retrained global classifier. The better the global classifier generalizes, the more it forces the local feature extractor to learn more generalized features. The local feature extractor's parameters are updated based on the gradient of the new loss:
  \begin{equation}
  	\Delta f_i^{\text{new}} = \nabla_{f_i} \mathcal{L}_i^{\text{new}}.
  \end{equation}
  
  This means that the feature extractor is adjusting to produce features that are more aligned with the improved, generalized classifier. Over time, this feedback loop helps the feature extractor produce features that generalize better. This process of improving the feature extractor based on the retrained global classifier happens iteratively over several rounds of local training. As each local feature extractor is guided by a more generalized global classifier, the feature extractor continues to adjust and improve. This process makes the feature extractor's outputs $ \mathbf{z}_i $ progressively more generalized and suitable for a wide range of data distributions. Specifically, we have:
 \begin{equation}
 	\|\mathbf{z}_i^{\text{new}}\| \geq \|\mathbf{z}_i^{\text{previous}}\|,
 \end{equation}
  where  $ \mathbf{z}_i^{\text{new}} $ represents the updated feature vector after the feature extractor has been guided by the improved classifier, and $ \mathbf{z}_i^{\text{previous}} $ represents the feature vector before retraining.
   \texttt{FedAvg} aggregates local models from each client and sends them to the server, where a global classifier is trained on aggregated features. The global classifier is retrained and improved, gaining a broader understanding of the entire data distribution. The improved global classifier is then used to guide the local feature extractors in the next local round, encouraging them to produce more generalized feature representations. Over multiple local epochs, this process repeats, allowing the feature extractors to gradually produce better, more generalized features.
\end{proof}
\begin{table*}[!t]
	\centering
	\caption{Backbone Architecture}
	\begin{tabular}{p{3cm} p{5cm} p{5cm} }
		\hline
		\textbf{Layer}    & \textbf{MLP}        & \textbf{CNN}           \\ \hline
		Feature Extractor 1    & input: (flatten image), 512  & input: 1, output: 32, kernel: 5 \\ 
		Feature Extractor 2    & 512, 512            & input: 32, output: 64, kernel: 5 \\ 
		Classifier 1       & 512, Number of classes    & output features, 512        \\ 
		Classifier 2       & ---                 & 512, Number of classes             \\ \hline
	\end{tabular}
	\label{arc}
\end{table*}

\section{Complexity Analysis for FedFeat+}
In this section, we analyze the computational complexity of our proposed FedFeat+ algorithm and compare it with the standard FedAvg algorithm.

\textit{Local Training Complexity (Client Side):}
Each client trains its local model on its dataset for a given number of epochs. If the dataset size for client $i$ is denoted as $N_i$ and the number of training epochs is $E$, the complexity of the local training for each client is:
\[ \mathcal{O}(N_i \cdot E \cdot D), \]
where $D$ is the number of parameters in the model.

\textit{Feature Perturbation Complexity (Client Side):}
The perturbation step involves adding noise to the features before sending them to the server. Assuming the number of features is $F$ for each client, the complexity of perturbing the features for each client is:
\[
	\mathcal{O}(F).
\]

\textit{Server Aggregation Complexity:}
The server aggregates the models by averaging the weights from each client. If there are $N$ clients, the aggregation step involves summing the weights and averaging them. The complexity of this step is:
\[
	\mathcal{O}(N \cdot D),
\]
where $N$ is the number of clients and $D$ is the number of parameters in the model.

\textit{Retraining Complexity (Server Side):}
After aggregation, the server retrains the global model using the aggregated and perturbed features. Let the number of aggregated features be $N \times F$ (i.e., the total number of features from all clients), and the number of epochs for retraining is $E_r$. The complexity of retraining the global classifier is:
\[
	\mathcal{O}(N \cdot F \cdot E_r \cdot D).
\]

\textit{Total Complexity per Global Round:}
The total complexity of each global round (including local training, feature perturbation, server aggregation, and retraining) is the sum of the individual complexities:
\[
	\mathcal{O}\left( \sum_{i=1}^{N} (N_i \cdot E \cdot D) + N \cdot F + N \cdot D + N \cdot F \cdot E_r \cdot D \right).
\]

Simplifying the terms we get:
\[
	\mathcal{O}\left( N \cdot \left( N_{\text{max}} \cdot E \cdot D + F + D + F \cdot E_r \cdot D \right) \right),
\]
where $N_{\text{max}}$ is the maximum number of data points across all clients.

\textit{Comparison with FedAvg:}
In FedAvg, without the perturbation and retraining steps, the complexity mainly comes from local training and server aggregation:
\[
	\mathcal{O}(N \cdot N_{\text{max}} \cdot E \cdot D + N \cdot D).
\]

The key difference in FedFeat+ is the addition of the perturbation and retraining steps, which introduce extra complexity. However, the complexity of perturbation (\(\mathcal{O}(F)\)) and retraining (\(\mathcal{O}(N \cdot F \cdot E_r \cdot D)\)) are relatively small compared to the local training step (\(\mathcal{O}(N_{\text{max}} \cdot E \cdot D)\)).

The extra steps in FedFeat+ (perturbation and retraining) increase the overall complexity, but the impact on performance is not significant because the number of features \(F\) is typically much smaller than the dataset size or model parameters, making the perturbation step computationally inexpensive. The retraining is performed on the aggregated features rather than the full dataset, reducing the computational load. Moreover, the retraining epochs \(E_r\) are typically kept small. Even with the additional complexity, FedFeat+ remains scalable because the added steps only introduce a small overhead compared to the primary computational cost of local training, which dominates the overall complexity in large-scale FL systems. Although FedFeat+ introduces additional computational steps compared to FedAvg, these steps are not prohibitively expensive and the algorithm remains computationally efficient, especially when considering its improved accuracy and privacy-preserving features. Therefore, the additional complexity is a reasonable trade-off for the benefits it provides.

\section{Experiment} \label{sec6} \label{Exp}
\subsection{Dataset and Implemental Details}
In our research, we employed four standard datasets: CIFAR10 \cite{cifar}, CIFAR100 \cite{cifar}, MNIST \cite{mnist}, and Fashion MNIST \cite{fmnist}. Experiments were conducted under two client configurations: 10 clients and 50 clients. For the IID distribution, data was proportionally split among clients, ensuring a balanced representation of all classes. For the non-IID distribution, we used a semi-pathological partitioning method, sorting data by class labels and dividing it into shards, each containing 300 examples. Each client was randomly assigned two shards, resulting in most clients having examples from only two or fewer classes. This approach introduced significant data heterogeneity, creating a challenging scenario to evaluate model performance under non-IID conditions.

The proposed model was implemented using Python \cite{Python} 3.10.12 and PyTorch \cite{pytorch} 2.0. We employed the Adam optimizer with a learning rate of 0.01 and trained for five local epochs per client. Depending on hardware constraints, we performed between 100 and 300 communication rounds to comprehensively assess performance across experimental setups. The chosen hyperparameters for the DP mechanism are $\epsilon = 2.0$ for CIFAR-10 and CIFAR-100, and $\epsilon = 1.5$ for MNIST and Fashion-MNIST, with a constant sensitivity $\Delta F = 1.0$ across all datasets. For evaluation, we utilized two backbone architectures: a multi-layer perceptron (MLP) and a convolutional neural network (CNN). To ensure computational efficiency, we opted for lightweight backbone models. The architecture details of these backbones are outlined in TABLE \ref{arc}, designed to balance model complexity and computational feasibility.

To evaluate the effectiveness of DP noise, we employ four key metrics: mutual information (MI), entropy difference (ED), Kullback-Leibler (KL) divergence, and feature correlation (FC). MI is used to analyze privacy leakage and assess the effectiveness of noise addition. A high MI after adding DP noise indicates that sensitive information remains in the output, posing a privacy risk. Conversely, a low MI suggests that the DP noise effectively reduces the correlation between the original and perturbed features, thereby enhancing privacy. \emph{Mutal Information} quantifies the shared information between two distributions and is defined as follows:
\begin{equation}
	MI(X; Y) = H(X) + H(Y) - H(X, Y),
\end{equation}
where $H(X)$ is the entropy of the original feature distribution, $H(Y)$ is the entropy of the noisy feature distribution, $H(X, Y)$  is the joint entropy of $X$ and $Y$. \emph{Entropy} is used to measure the effect of noise and the extent of information loss. The addition of DP noise influences the randomness of the feature distribution. A significant decrease in entropy may indicate that the noise is overly dominant, reducing feature diversity. A large entropy drop before and after noise addition suggests substantial information loss, which could negatively impact model accuracy. Entropy quantifies the uncertainty of a probability distribution and is defined as follows:
\begin{equation}
	ET(X) = - \sum_{i} p_i \log (p_i + \epsilon_1),
\end{equation}
where $X$ is a random variable representing the feature distribution, $p_i$  is the probability of the $i$-th feature value, $\epsilon_1$ is a small constant (e.g., $10^{-9}$) for numerical stability. \emph{Entropy difference} quantifies the change in entropy between the original and noisy features, measuring the impact of noise on feature uncertainty. It is defined as follows:
\begin{equation}
	ED = |ET(X) - ET(\widetilde{X})|,
\end{equation}
where $X$ and $\widetilde{X}$ represent the fresh and noisy features respectively. \emph{KL divergence} quantifies the shift between the original and noisy feature distributions, serving as an indicator of the privacy-utility trade-off. A high KL divergence suggests that the added DP noise has significantly altered the distribution, potentially impacting model performance. Conversely, a very low KL divergence may indicate insufficient noise, failing to ensure privacy. Striking a balance is crucial to preserving both privacy and utility. KL divergence, which measures how one probability distribution diverges from another, is defined as follows:
\begin{equation}
	D_{KL}(P || Q) = \sum_{i} p_i \log \left( \frac{p_i + \epsilon_2}{q_i + \epsilon_2} \right),
\end{equation}
where, $P$ is the probability distribution before DP noise, $Q$ is the probability distribution after DP noise, $p_i$ and $q_i$ are the probabilities of feature values before and after noise, $\epsilon_2$ prevents division by zero. Feature correlation evaluates the preservation of feature structure and helps detect privacy risks. If correlation remains high after adding DP noise, the feature relationships are well-preserved, maintaining model utility. However, a high correlation may also indicate that the noise is insufficient to obscure feature values, potentially exposing sensitive information. Striking a balance between privacy and utility is essential. \emph{Feature correlation}, which quantifies the linear dependence between original and noisy features, is defined as follows:
\begin{equation}
	FC(X,Y) = \frac{\mathbb{E}[(X - \mu_X)(Y - \mu_Y)]}{\sigma_X \sigma_Y},
\end{equation}
where, $\mathbb{E}[.]$  denotes expectation, $\mu_X$, $\mu_Y$  are the means of the original and noisy features, $\sigma_X, \sigma_Y$  are the standard deviations of $X$ and $Y$.

\begin{figure*}[!t]
	\centering
	\includegraphics[height = 6cm, width=.48\linewidth]{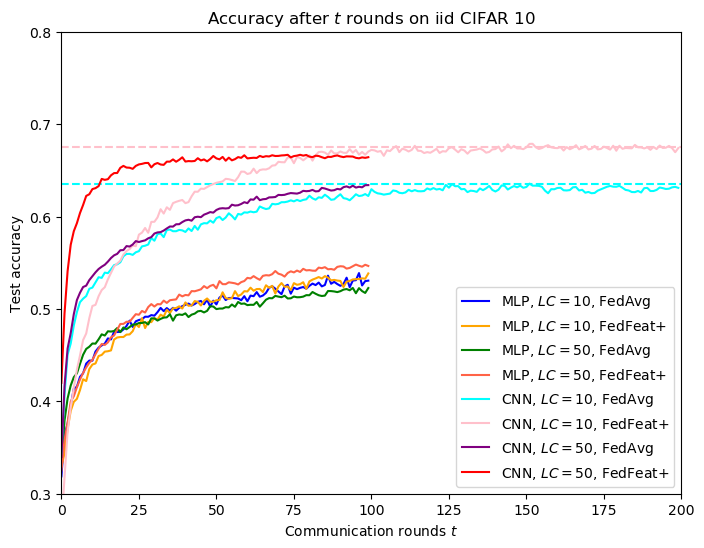}
	\includegraphics[height = 6cm, width=.48\linewidth]{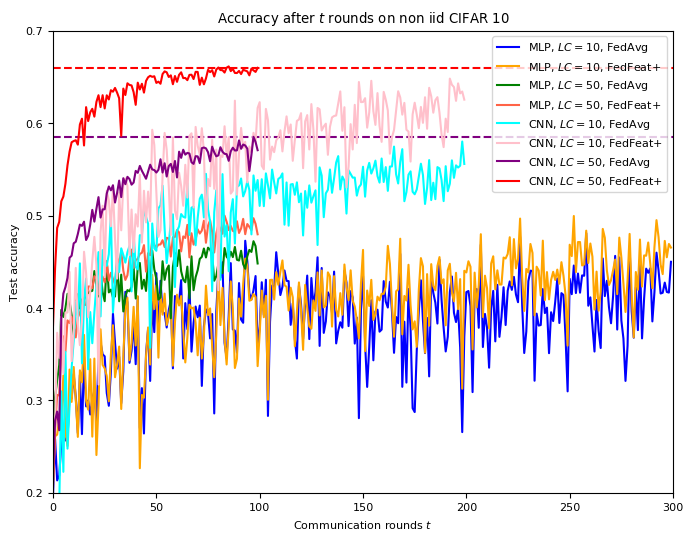}\\
	\centering (a) iid scenario	\hspace{50mm} (b) non iid scenario\\
	\caption{Comparison of communication rounds and accuracy on CIFAR-10 considering both iid and non-iid data distribution.}
	\label{c10}
\end{figure*}
\begin{figure*}[!t]
	\centering
	\includegraphics[height = 6cm, width=.48\linewidth]{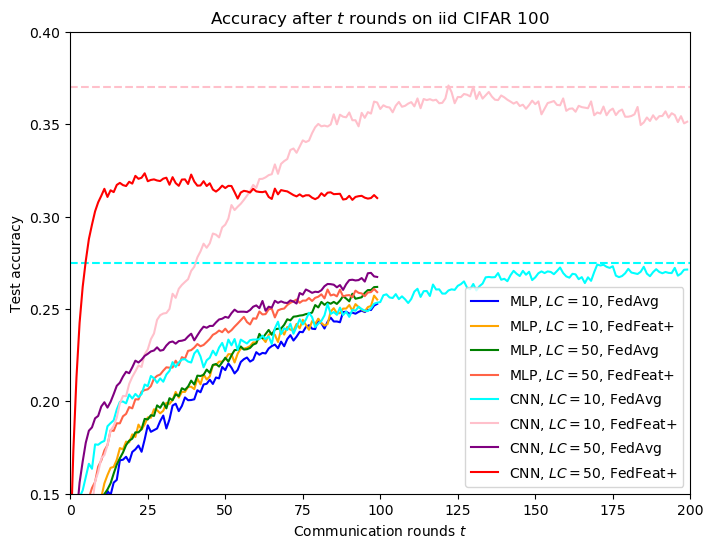}
	\includegraphics[height = 6cm, width=.48\linewidth]{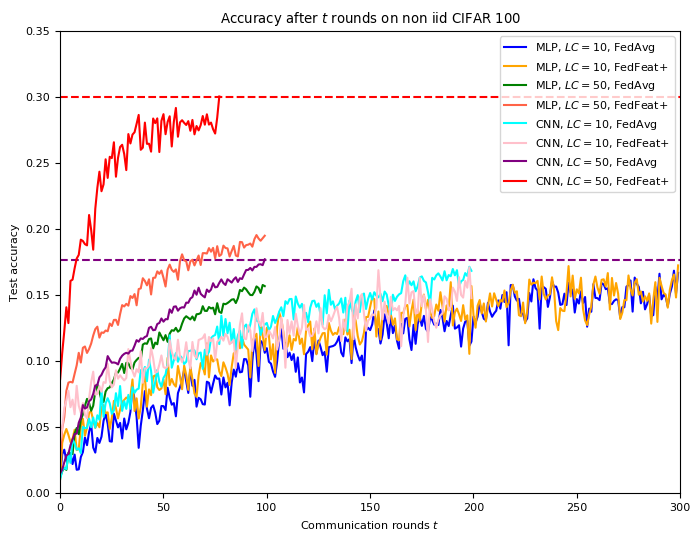}\\
	\centering (a) iid scenario \hspace{50mm} (b) non iid scenario\\
	\caption{Comparison of communication rounds and accuracyy on CIFAR-100 considering both iid and non-iid data distribution.}
	\label{c100}
\end{figure*}

\subsection{Results}
In this section, we evaluate our model's performance using accuracy as the primary metric. Fig. \ref{c10} presents the accuracy and communication efficiency of the proposed FedFeat+ method compared to FedAvg on the CIFAR-10 dataset under both IID and non-IID data distributions. In Fig. \ref{c10}(a), we show the results for the IID case. For the MLP model, the accuracy of FedFeat+ and FedAvg remains similar. However, in the case of CNN, FedFeat+ consistently outperforms FedAvg for both 10 and 50 local clients. These results indicate that FedFeat+ not only achieves higher accuracy but also converges significantly faster, especially with the CNN backbone. In the non-IID setting, shown in Fig. \ref{c10}(b), FedFeat+ demonstrates faster convergence and higher accuracy. For the MLP model, FedFeat+ achieves better accuracy and converges faster than FedAvg. A similar trend is observed in the CNN model, where FedFeat+ attains superior accuracy and faster convergence than FedAvg. Therefore, it is evident that the proposed FedFeat+ can handle extreme imbalance data across diverse FL environments.

In Fig. \ref{c100}, we compare the accuracy and convergence efficiency of FedFeat+ and FedAvg on the CIFAR-100 dataset under both IID and non-IID data distributions. Unlike CIFAR-10, CIFAR-100 presents a significantly greater challenge due to its increased complexity, featuring the same images but with ten times more classes. In Fig. \ref{c100}(a), we present the IID case. For the MLP model, the accuracy of FedFeat+ and FedAvg remains similar. However, for the CNN model, FedFeat+ achieves significantly higher accuracy than FedAvg for both 10 and 50 local clients. In the non-IID scenario, shown in Fig. \ref{c100}(b), we observe that for 10 local clients, both FedFeat+ and FedAvg converge similarly and provide comparable accuracies for both MLP and CNN models. However, in the 50 local client setting, FedFeat+ converges much faster and achieves significantly higher accuracy than FedAvg for both models. These results indicate that FedAvg struggles on this dataset, primarily due to the simplicity of the backbone models used. While FedFeat+ shows limited improvement with 10 local clients, it achieves a notable enhancement in convergence and accuracy when the number of clients increases to 50. This improvement is particularly pronounced in the non-IID setting, where FedFeat+ outperforms FedAvg, demonstrating its effectiveness in handling complex and imbalanced data distributions.

\begin{figure*}[!t]
	\centering
	\includegraphics[height = 6cm, width=.48\linewidth]{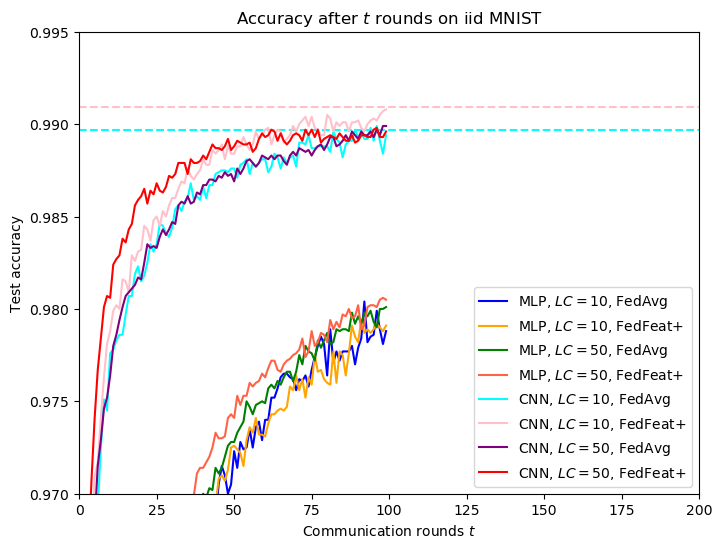}
	\includegraphics[height = 6cm, width=.48\linewidth]{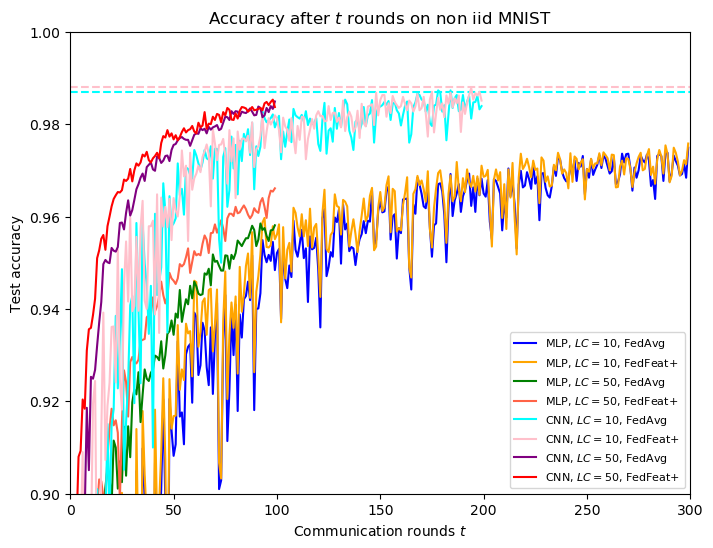}\\
	\centering (a) iid scenario \hspace{50mm} (b) non iid scenario\\
	\caption{Comparison of communication rounds and accuracy on MNIST considering both iid and non-iid data distribution.}
	\label{mnist}
\end{figure*}

Fig. \ref{mnist} illustrates the accuracy and convergence efficiency of FedFeat+ in comparison to FedAvg on the MNIST dataset, considering both IID and non-IID data distributions. Given the simplicity of the MNIST dataset, FedAvg demonstrates a high level of performance across different backbone architectures, achieving near-perfect accuracy. While the potential for substantial accuracy improvements is limited within this context, it is noteworthy that FedFeat+ maintains comparable levels of accuracy without significant degradation. Moreover, FedFeat+ exhibits a slight performance enhancement over FedAvg in both IID and non-IID scenarios, demonstrating its potential for achieving competitive results even on datasets where traditional methods like FedAvg already exhibit strong performance.
\begin{figure*}[!t]
	\centering
	\includegraphics[height = 6cm, width=.48\linewidth]{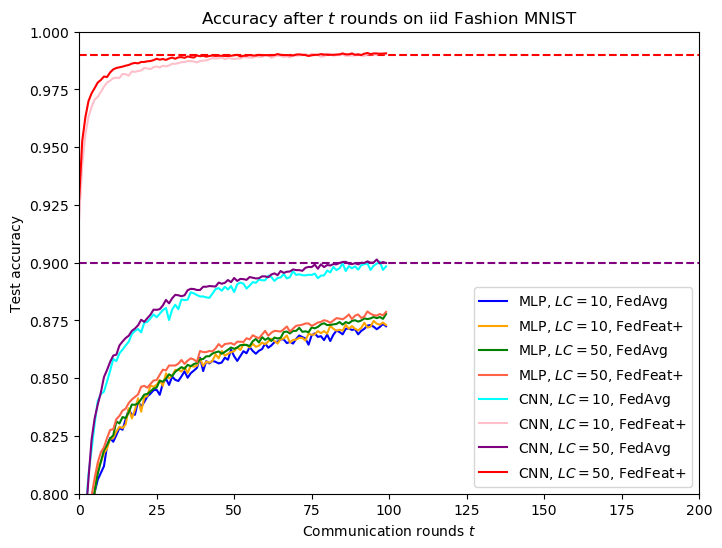}
	\includegraphics[height = 6cm, width=.48\linewidth]{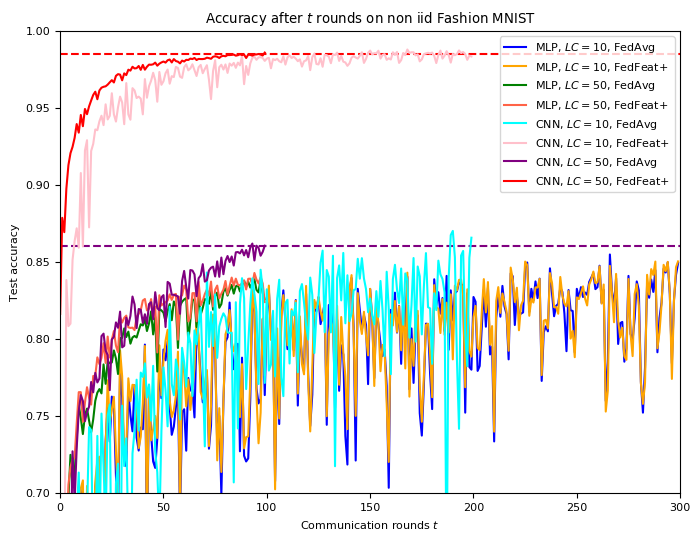}\\
	\centering (a) iid scenario \hspace{50mm} (b) non iid scenario\\
	\caption{Comparison of communication rounds and accuracy on Fashin MNIST considering both iid and non-iid data distribution.}
	\label{fmnist}
\end{figure*}

Fig. \ref{fmnist} presents a comparative analysis of the accuracy and convergence efficiency of FedFeat+ and FedAvg on the Fashion MNIST dataset, a more challenging benchmark than the standard MNIST dataset. In Fig. \ref{fmnist}(a), we present the IID scenario. For the MLP model, both FedFeat+ and FedAvg exhibit similar convergence rates and accuracies across 10 and 50 local clients. However, for the CNN model, FedFeat+ converges significantly faster and achieves notably higher accuracy for both client settings. These results indicate that while FedAvg performs competitively across different architectures, FedFeat+ consistently outperforms it in accuracy. In Fig. \ref{fmnist}(b), we present the non-IID scenario. Similar to the IID case, the MLP model shows comparable accuracy and convergence for both methods across 10 and 50 local clients. However, for the CNN model, FedFeat+ demonstrates significantly superior convergence speed and accuracy compared to FedAvg in both client settings. In this non-IID setting, where data is unevenly distributed among devices, FedFeat+ exhibits greater stability and robustness. Its convergence trajectory is smoother and less prone to oscillations compared to FedAvg, indicating a more effective and resilient approach to handling data heterogeneity.

In TABLE \ref{iid_table}, we present the maximum accuracy achieved by the FedFeat+ method compared to FedAvg, using both MLP and CNN backbone architectures under IID settings with 10 and 50 local clients. We evaluate the performance across four datasets, where LC denotes the number of local epochs. Our results show that FedFeat+ consistently outperforms FedAvg in nearly all configurations, especially on more complex datasets such as CIFAR-10, CIFAR-100, and Fashion MNIST. We observe that the performance gap is significantly larger with the CNN backbone, which demonstrates its superior ability to capture and utilize the intricate patterns present in complex datasets. These findings confirm the effectiveness of FedFeat+ in FL scenarios. In TABLE \ref{noniid_table}, we extend the analysis conducted in TABLE \ref{iid_table} to explore the performance of FedFeat+ and FedAvg under non-IID settings. This comparison provides deeper insights into how the methods perform when data distributions across clients are highly heterogeneous. Our results clearly demonstrate that FedFeat+ consistently surpasses FedAvg in all scenarios. Notably, the performance gap between the two methods is significantly larger in the non-IID context than in the IID scenario, further emphasizing the adaptability of FedFeat+ to real-world FL challenges. We observe that this enhanced performance stems from FedFeat+’s ability to effectively handle the complexities and inconsistencies of non-IID data distributions. By leveraging its superior feature representation capabilities, FedFeat+ mitigates the adverse effects of client heterogeneity, which often hinder the performance of conventional methods like FedAvg. 

\begin{table}[t]
	\centering
	\caption{Accuracy comparison between the proposed method and FedAvg across different datasets with IID data distribution.}
	\begin{tabular}{ccccccc}
		\hline
		\multirow{2}{*}{\textbf{Model}} & \multirow{2}{*}{\textbf{Method}} & \multirow{2}{*}{\textbf{LC}} & \multicolumn{4}{c}{\textbf{Maximum Accuracy}} \\ \cline{4-7} 
		& & & cifar10 & cifar100 & mnist & fmnist \\ \hline
		\multirow{4}{*}  & FedAvg  & 10 & 53.84 & 25.27 & 98.04 & 87.35 \\ 
		& FedFeat+  & 10 & 53.88 & 25.72 & 97.94 & 87.48 \\ 
		MLP& FedAvg  & 50 & 52.27 & 26.08 & 98.01 & 87.78 \\ 
		& FedFeat+  & 50 & \textbf{54.80} & \textbf{26.19} & \textbf{98.06} & \textbf{87.89} \\ \hline
		\multirow{4}{*} & FedAvg  & 10 & 63.95 & 27.62 & \textbf{98.99} & 90.05 \\ 
		& FedFeat+  & 10 & \textbf{67.87} & \textbf{37.09} & 99.08 & 99.07 \\ 
		CNN& FedAvg  & 50 & 63.40 & 26.95 & \textbf{98.99} & 90.14 \\ 
		& FedFeat+  & 50 & 66.68 & 32.34 & 98.98 & \textbf{99.08} \\ \hline
	\end{tabular}
	\label{iid_table}
\end{table}

\begin{table}[t]
	\centering
	\caption{Accuracy comparison between the proposed method and FedAvg across different datasets with Non IID data distribution.}
	\begin{tabular}{ccccccc}
		\hline
		\multirow{2}{*}{\textbf{Model}} & \multirow{2}{*}{\textbf{Method}} & \multirow{2}{*}{\textbf{LC}} & \multicolumn{4}{c}{\textbf{Maximum Accuracy}} \\ \cline{4-7} 
		& & & cifar10 & cifar100 & mnist & fmnist \\ \hline
		\multirow{4}{*}  & FedAvg  & 10 & 48.09 & 16.82 & 97.49 & 85.47 \\ 
		& FedFeat+  & 10 & 49.95 & 17.22 & 97.58 & 85.02 \\ 
		MLP& FedAvg  & 50 & 49.94 & 19.91 & 97.61 & 86.15 \\ 
		& FedFeat+  & 50 & \textbf{50.96} & \textbf{22.29} & \textbf{97.72} & \textbf{86.33} \\ \hline
		\multirow{4}{*} & FedAvg  & 10 & 58.01 & 17.12 & 98.73 & 87.00 \\ 
		& FedFeat+  & 10 & 64.82 & 17.08 & \textbf{98.78} & \textbf{98.75} \\ 
		CNN& FedAvg  & 50 & 58.52 & 17.68 & 98.49 & 86.18 \\ 
		& FedFeat+  & 50 & \textbf{66.14} & \textbf{30.02} & 98.53 & 98.58 \\ \hline
	\end{tabular}
	\label{noniid_table}
\end{table}

From our experimental analysis, we observe that both FedAvg and FedFeat+ exhibit relatively smooth convergence under IID settings. However, in non-IID scenarios, FedAvg struggles with gradient stability, resulting in erratic convergence characterized by a pronounced zigzag pattern. In contrast, FedFeat+ achieves significantly smoother convergence, especially when leveraging a CNN backbone. These results highlight the capability of our proposed FedFeat+ framework to effectively address the challenges posed by hostile and highly imbalanced data distributions. Its ability to maintain gradient stability and achieve consistent convergence, even under extreme imbalance, underscores its robustness. This stability positions FedFeat+ as a practical and reliable solution for real-world FL applications where data heterogeneity is a critical concern. The greater performance gap observed with the CNN architecture compared to the MLP can be attributed to the presence of dual classification layers in the CNN. These layers play a crucial role in enhancing the model's capacity to adapt to diverse data distributions and extract more intricate feature representations. By utilizing dual classification layers, the CNN can refine its decision boundaries more effectively, resulting in higher accuracy compared to the single classification layer in the MLP. This observation strongly supports our primary proposal that classifier retraining is a critical factor in boosting performance. Models equipped with multiple classification layers, like CNNs, are inherently better suited to exploit the benefits of retraining due to their ability to process and adapt to complex patterns within the data.
\begin{table}[t]
	\centering
	\caption{Accuracy vs. privacy evaluation for different privacy budgets ($\epsilon$) with fixed sensitivity on CIFAR-10 using our proposed method in a IID setting across 10 local clients.}
	\begin{tabular}{c c c c c c c c}
		\hline
		\textbf{Model} & $\epsilon$ & $\Delta$F &  Acc.$\uparrow$ & $MI\downarrow$ & $ED\uparrow$ & $FC\downarrow$ & $D_{KL}\uparrow$ \\ \hline
		\multirow{4}{*} {MLP}& 0.5 & 1.0  &  44.57 & \textbf{0.0099} & \textbf{0.0051} & \textbf{0.0002} & \textbf{0.0030} \\
		& 1.0  & 1.0  & 46.35  & 0.0112 & 0.0043 & 0.0003 & 0.0022 \\
		& 1.5  & 1.0  & 49.65 & 0.0121 & 0.0032 & 0.0004 & 0.0011 \\ 
		& 2.0  & 1.0  & \textbf{53.88}  & 0.0129 & 0.0021 & 0.0005 & 0.0008 \\ \hline
		\multirow{4}{*} {CNN} & 0.5  & 1.0  & 62.36  & \textbf{0.0171} & \textbf{0.0007} & \textbf{0.0013} & \textbf{0.0011} \\ 
		& 1.0  & 1.0  & 64.03  & 0.0178 & 0.0006 & 0.0014 & 0.0009 \\
		& 1.5  & 1.0  &  65.71  & 0.0182 & 0.0005 & 0.0015 & 0.0006 \\
		& 2.0  & 1.0  &  \textbf{67.87} & 0.0185 & 0.0004 & 0.0016 & 0.0004 \\ \hline
	\end{tabular}
	\label{iid_hp}
\end{table}

To evaluate the trade-off between privacy and utility in our proposed method, we conducted hyperparameter testing by varying the privacy budget (\(\epsilon\)) while keeping the sensitivity (\(\Delta F\)) fixed at 1.0. Our evaluation was performed on the CIFAR-10 dataset in an IID setting across 10 local clients, using both MLP and CNN models. As shown in TABLE \ref{iid_hp}, we observe that increasing \(\epsilon\) leads to a gradual improvement in accuracy, as the amount of added noise decreases, thereby allowing the classifier to make better predictions. However, excessive reduction in noise compromises privacy, as reflected in the MI and FC values. We note that for lower privacy budgets (\(\epsilon = 0.5\)), MI and FC values are minimized, indicating stronger privacy guarantees, but at the cost of lower accuracy. Conversely, at higher values (\(\epsilon = 2.0\)), accuracy peaks at 53.88\% (MLP) and 67.87\% (CNN), while MI, ED, FC, and KL divergence (\(D_{KL}\)) remain within acceptable limits. Based on these observations, we select \(\epsilon = 2.0\), \(\Delta F = 1.0\) as the suboptimal configuration, achieving a balanced trade-off between privacy preservation and classification performance.
\begin{table}[t]
	\centering
	\caption{Accuracy vs. privacy evaluation for different privacy budgets ($\epsilon$) with fixed sensitivity on CIFAR-10 using our proposed method in a non-IID setting across 10 local clients.}
	\begin{tabular}{c c c c c c c c}
		\hline
		\textbf{Model} & $\epsilon$ & $\Delta$F &  Acc.$\uparrow$ & $MI\downarrow$ & $ED\uparrow$ & $FC\downarrow$ & $D_{KL}\uparrow$ \\ \hline
		\multirow{4}{*} {MLP}& 0.5 & 1.0  & 36.76 & \textbf{0.0109} & \textbf{0.0049} & \textbf{0.0002} & \textbf{0.0030} \\
		& 1.0  & 1.0  & 40.98  & 0.0117 & 0.0043 & 0.0003 &0.0022 \\ 
		& 1.5  & 1.0  & 45.55 & 0.0122 & 0.0032 & 0.0004 & 0.0013 \\ 
		& 2.0  & 1.0  &  \textbf{49.95} & 0.0129 & 0.0021 & 0.0005 & 0.0009 \\ \hline
		\multirow{4}{*} {CNN} & 0.5  & 1.0  & 57.97  & \textbf{0.0196} & \textbf{0.0009} & \textbf{0.0007} & \textbf{0.0012} \\  
		& 1.0  & 1.0  & 59.04  & 0.0201 & 0.0008 & 0.0008 &0.0009 \\
		& 1.5  & 1.0  &  62.87  & 0.0212 & 0.0007 & 0.0009 & 0.0006 \\ 
		& 2.0  & 1.0  &  \textbf{64.82} & 0.0214 & 0.0005 & 0.0010 & 0.0005 \\ \hline
	\end{tabular}
	\label{noniid_hp}
\end{table}

In the case of non IID CIFAR 10 dataset, TABLE \ref{noniid_hp} presents the accuracy and privacy metric evaluations for different values of $\epsilon$ with a fixed sensitivity ($\Delta F = 1.0$). As expected, increasing $\epsilon$ leads to improved model utility (higher accuracy) at the cost of reduced privacy. For both MLP and CNN models, accuracy improves as $\epsilon$ increases, with the CNN model achieving the highest accuracy of 64.82\% at $\epsilon = 2.0$. However, privacy degradation is reflected in the increasing values of MI and KL divergence ($D_{KL}$), which indicate a higher correlation between original and perturbed features. To balance accuracy and privacy, we select $\epsilon = 2.0$ as the optimal choice, which provides satisfactory performance while maintaining reasonable privacy preservation. Compared to lower $\epsilon$ values, this setting retains acceptable MI, ED, FC, and KL divergence levels, ensuring that noise addition does not excessively distort the feature representations needed for classification.

\section{Conclusion} \label{sec7} \label{Con}
In this research, we propose the FedFeat+ framework, integrating FL with differential privacy mechanisms. By separating feature extraction from classification, we propose a robust two-tiered training process that enhances model accuracy while ensuring data privacy. Our convergence analysis confirms the framework's stability and ability to reach optimal solutions despite noise interference. Through empirical evaluations on datasets such as CIFAR-10, CIFAR-100, MNIST, and FMNIST, we demonstrated that the FedFeat+ framework outperforms traditional FedAvg methods. The experimental results reveal that the FedFeat+ framework significantly enhances performance compared to the FedAvg method, achieving improvements of approximately 4\%, 10\%, and 9\% on CIFAR-10, CIFAR-100, and Fashion MNIST datasets respectively, under IID scenarios, and 8\%, 13\%, and 12\% respectively under non-IID conditions, all utilizing a two-layer classifier backbone. These findings highlight that the FedFeat+ framework effectively enhances model generalization and captures the holistic data distribution, thereby successfully addressing the accuracy and privacy challenges associated with extreme non-IID conditions and highly imbalanced data distributions in FL applications, particularly in IoT environments where device heterogeneity and privacy concerns are critical.

%
%
%
%
%
%
%

\bibliographystyle{IEEEtran}
\bibliography{references.bib}

%
%

\begin{IEEEbiography}[{\includegraphics[width=1in,height=1.2in,clip]{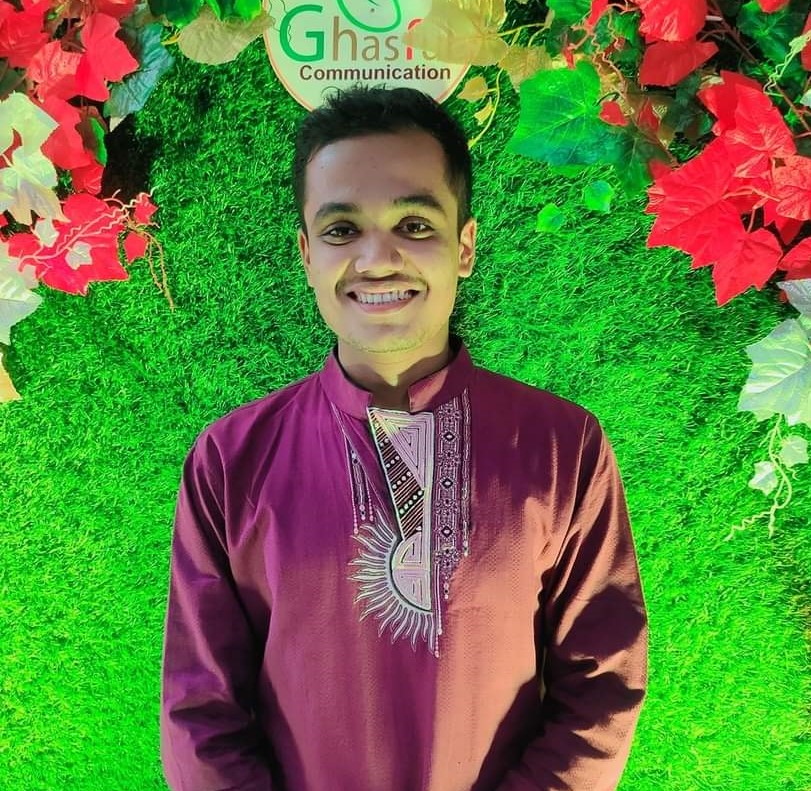}}]{Mrityunjoy Gain}
Mrityunjoy Gain received the B.S. degree in computer science from Khulna University, Bangladesh, in 2021. Currently, he is doing the M.S. leading PhD in Artificial Intelligence at Kyung Hee University, Korea, and working in the Networking Intelligence Lab. His research interests includes computer vision, continual learning, deep learning, open RAN, 6G, and pattern recognition.
\end{IEEEbiography}

\begin{IEEEbiography}[{\includegraphics[width=1in,height=1.5in,clip,keepaspectratio]{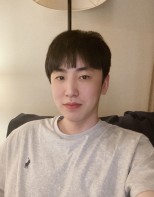}}]{Kitae Kim}
	received his B.S., M.S., and Ph.D. degrees in computer science and engineering from Kyung Hee University, Seoul, South Korea, in 2017, 2019, and 2024, respectively. He is currently working as a Postdoctoral Researcher in the Networking Intelligence Laboratory at Kyung Hee University, Seoul, South Korea. His research interests include 5G/6G wireless communication, channel estimation, channel prediction, and machine learning.
\end{IEEEbiography}

\begin{IEEEbiography}[{\includegraphics[width=1in,height=1.5in,clip,keepaspectratio]{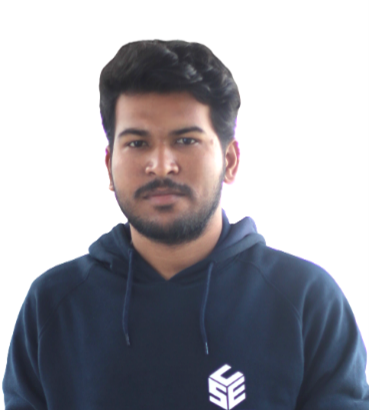}}]{Avi Deb Raha}
received the B.S. degree in computer science from Khulna University, Bangladesh, in 2020. Currently he is a PhD student at the Department of Computer Science and Engineering at Kyung Hee University, South Korea. His research interests are currently focused on Semantic Communication, Deep Learning, Generative AI, holographic MIMO, and Integrated Sensing and Communication. 
\end{IEEEbiography}

\begin{IEEEbiography}[{\includegraphics[width=1in,height=1.5in,clip,keepaspectratio]{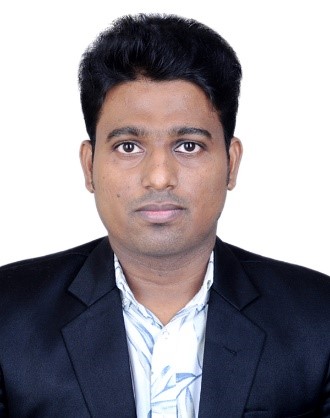}}]{Apurba Adhikary} received his B.Sc and M.Sc Engineering degrees in Electronics and Communication Engineering from Khulna University, Khulna, Bangladesh in 2014 and 2017, respectively. He is a Ph.D. Researcher in the Department of Computer Science and Engineering at Kyung Hee University (KHU), South Korea. He has been serving as an Assistant Professor in the Department of Information and Communication Engineering at Noakhali Science and Technology University (NSTU), Noakhali, Bangladesh since 28 January 2020. In addition, he served as a Lecturer in the Department of Information and Communication Engineering at Noakhali Science and Technology University (NSTU), Noakhali, Bangladesh from 28 January 2018 to 27 January 2020. His research interests are currently focused on integrated sensing and communication, holographic MIMO, cell-free MIMO, intelligent networking resource management, artificial intelligence, and machine learning. He received the Best Paper Award at the 2023 International Conference on Advanced Technologies for Communications (ATC) in 2023.
\end{IEEEbiography}

\begin{IEEEbiography}[{\includegraphics[width=1in,height=1.5in,clip,keepaspectratio]{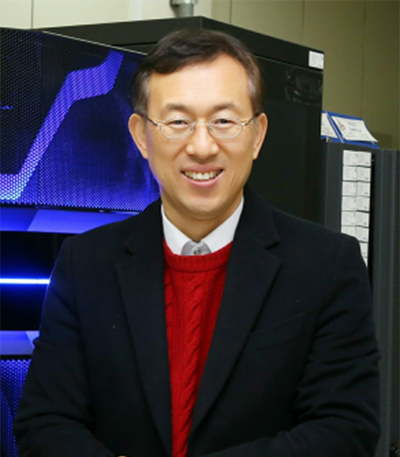}}]{Eui-Nam Huh}
	(Member, IEEE) received the B.S. degree from Busan National University, South Korea, the master’s degree in computer science from The University of Texas at Arlington, USA, in 1995, and the Ph.D. degree from The Ohio University, USA, in March, 2002. Currently, he is a Professor with the Department of Computer Science and Engineering, Kyung Hee University, South Korea. His research interests include a diverse range of subjects, such as AI for cloud computing, Cloud for AI, AIoT, AI for network, parallel \& federated learning, mobile computing, big data, and security. He served on the review board for the National Research Foundation of Korea. He has actively participated in community services for several organizations, including Applied sciences, KSII, ITU-T SG13, WPDRTS/IPDPS, APAN Sensor Network Group, ICUIMC, ICONI, APICIST, and ICUFN as different types of chairs and editor.
\end{IEEEbiography}

\begin{IEEEbiography}[{\includegraphics[width=1in,height=1.5in,clip,keepaspectratio]{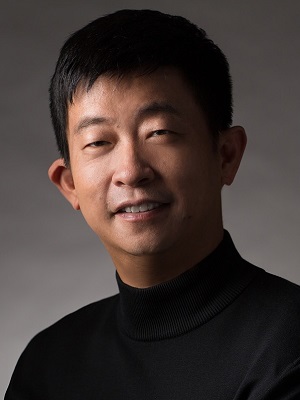}}]{Zhu Han}
	(S’01-M’04-SM’09-F’14) received the B.S. degree in electronic engineering from Tsinghua University, in 1997, and the M.S. and Ph.D. degrees in electrical and computer engineering from the University of Maryland, College Park, in 1999 and 2003, respectively. Currently, he is a John and Rebecca Moores Professor in the Electrical and Computer Engineering Department as well as in the Computer Science Department at the University of Houston, Texas. Dr. Han’s main research targets on the novel gametheory related concepts critical to enabling efficient and distributive use of wireless networks with limited resources. His other research interests include wireless resource allocation and management, wireless communications and networking, quantum computing, data science, smart grid, carbon neutralization, security and privacy. Dr. Han received an NSF Career Award in 2010, the Fred W. Ellersick Prize of the IEEE Communication Society in 2011, the EURASIP Best Paper Award for the Journal on Advances in Signal Processing in 2015, IEEE Leonard G. Abraham Prize in the field of Communications Systems (best paper award in IEEE JSAC) in 2016, IEEE Vehicular Technology Society 2022 Best Land Transportation Paper Award, and several best paper awards in IEEE conferences. Dr. Han was an IEEE Communications Society Distinguished Lecturer from 2015 to 2018 and ACM Distinguished Speaker from 2022 to 2025, AAAS fellow since 2019, and ACM Fellow since 2024. Dr. Han is a 1\% highly cited researcher since 2017 according to Web of Science. Dr. Han is also the winner of the 2021 IEEE Kiyo Tomiyasu Award (an IEEE Field Award), for outstanding early to mid-career contributions to technologies holding the promise of innovative applications, with the following citation: ``for contributions to game theory and distributed management of autonomous communication networks".
\end{IEEEbiography}

\begin{IEEEbiography}[{\includegraphics[width=1in,height=1.5in,clip,keepaspectratio]{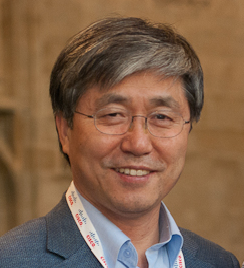}}]{Choong Seon Hong}(S’95-M’97-SM’11-F’24) received the B.S. and M.S. degrees in electronic engineering
	from Kyung Hee University, Seoul, South Korea, in 1983 and 1985, respectively, and the Ph.D. degree from Keio University, Tokyo, Japan, in 1997. In 1988, he joined KT, Gyeonggi-do, South Korea, where he was involved in broadband networks as a member of the Technical Staff. Since 1993, he has been with Keio University. He was with the Telecommunications Network Laboratory, KT, as a Senior Member of Technical Staff and as the Director of the Networking Research Team until 1999. Since 1999, he has been a Professor with the Department of Computer Science and Engineering, Kyung Hee University. His research interests include future Internet, intelligent edge computing, network management, and network security. Dr. Hong is a member of the Association for Computing Machinery (ACM), the Institute of Electronics, Information and Communication Engineers (IEICE), the Information Processing Society of Japan (IPSJ), the Korean Institute of Information Scientists and Engineers (KIISE), the Korean Institute of Communications and Information Sciences (KICS), the Korean Information Processing Society (KIPS), and the Open Standards and ICT Association (OSIA). He has served as the General Chair, the TPC Chair/Member, or an Organizing Committee Member of international conferences, such as the Network Operations and Management Symposium (NOMS), International Symposium on Integrated Network Management (IM), Asia-Pacific Network Operations and Management Symposium (APNOMS), End-to-End Monitoring Techniques and Services (E2EMON), IEEE Consumer Communications and Networking Conference (CCNC), Assurance in Distributed Systems and Networks (ADSN), International Conference on Parallel Processing (ICPP), Data Integration and Mining (DIM), World Conference on Information Security Applications (WISA), Broadband Convergence Network (BcN), Telecommunication Information Networking Architecture (TINA), International Symposium on Applications and the Internet (SAINT), and International Conference on Information Networking (ICOIN). He was an Associate Editor of the IEEE TRANSACTIONS ON NETWORK AND SERVICE MANAGEMENT and the IEEE JOURNAL OF COMMUNICATIONS AND NETWORKS and an Associate Editor for the International Journal of Network Management and an Associate Technical Editor of the IEEE Communications Magazine. He currently serves as an Associate Editor for the International Journal of Network Management and Future Internet Journal.
\end{IEEEbiography}




\end{document}